\documentclass[nohyperref]{article}

\usepackage{microtype}
\usepackage{graphicx}
\usepackage{caption}
\usepackage{subcaption}
\usepackage{booktabs} 

\usepackage{hyperref}

\usepackage[accepted]{icml2022}

\usepackage{amsmath}
\usepackage{amssymb}
\usepackage{mathtools}
\usepackage{amsthm}

\usepackage{amsfonts,bm}
\usepackage{amssymb,mathrsfs, amsthm}
\usepackage{xspace}
\usepackage{enumitem}
\usepackage{lipsum}
\usepackage{xpatch}
\usepackage{mathtools}
\usepackage{dsfont}
\usepackage{pgf,tikz}
\usetikzlibrary{positioning,matrix}
\usepackage{caption}
\usepackage{graphicx}

\def\mmkl{{\textsc{Meta-KeL}}\xspace}

\def\tH{{\gH_{k^*}}}
\def\Hhat{{\gH_{\hat{k}}}}
\def\teta{\eta^*}
\def\tveta{\bm{\eta}^*}
\def\tk{{k^*}}

\def\tvbeta{\bm{\beta}^*}
\def\khat{{\hat{k}}}

\def\etahat{\hat{\eta}}
\def\vbetahat{\hat{\bm{\beta}}}
\def\vetahat{\hat{\bm{\eta}}}
\def\gj{{^{(j)}}}
\def\grJ{{_{J_\tk}}}
\def\dmax{{d_{\mathrm{max}}}}
\def\kfull{{k_\mathrm{full}}}

\def\vvarepsilon{{\bm{\varepsilon}}}

\def\vbeta{{\bm{\beta}}}
\def\vphi{{\bm{\phi}}}
\def\veta{{\bm{\eta}}}

\def\vb{{\bm{b}}}

\def\vv{{\bm{v}}}

\def\vx{{\bm{x}}}
\def\vy{{\bm{y}}}

\def\mI{{\bm{I}}}

\def\mK{{\bm{K}}}

\def\mP{{\bm{P}}}

\def\mPhi{{\bm{\Phi}}}

\def\mPsi{{\bm{\Psi}}}

\def\gC{{\mathcal{C}}}
\def\gD{{\mathcal{D}}}

\def\gH{{\mathcal{H}}}

\def\gL{{\mathcal{L}}}

\def\gO{{\mathcal{O}}}

\def\gX{{\mathcal{X}}}

\def\sP{{\mathbb{P}}}

\def\sR{{\mathbb{R}}}

\newcommand{\tr}{\mathrm{Tr}}

\DeclareMathOperator*{\argmax}{arg\,max}

\DeclarePairedDelimiter\abs{\lvert}{\rvert}%
\DeclarePairedDelimiter\norm{\lVert}{\rVert}%

\makeatletter
\let\oldabs\abs
\def\abs{\@ifstar{\oldabs}{\oldabs*}}
\let\oldnorm\norm
\def\norm{\@ifstar{\oldnorm}{\oldnorm*}}
\makeatother

\makeatletter
\newcommand{\pushright}[1]{\ifmeasuring@#1\else\omit\hfill$\displaystyle#1$\fi\ignorespaces}
\newcommand{\pushleft}[1]{\ifmeasuring@#1\else\omit$\displaystyle#1$\hfill\fi\ignorespaces}
\makeatother

\newcommand{\vspaceexpfigure}{\vspace{-6pt}}
\newcommand{\vspaceparagraph}{\vspace{-2pt}}

\usetikzlibrary{fit}
\tikzset{%
  highlight/.style={rectangle,rounded corners,fill=comment,fill opacity=0.2,thick,inner sep=5pt}
}
\newcommand{\tikzmark}[2]{\tikz[overlay,remember picture,baseline=(#1.base)] \node (#1) {#2};}
\newcommand{\Highlight}[1][submatrix]{%
    \tikz[overlay,remember picture]{
    \node[highlight,fit=(left.north west) (right.east)] (#1) {};}
}
\tikzset{DG/.style =   {opacity=.2,line width=17 pt,line cap=round,color=comment}}
\tikzset{DGG/.style =  {opacity=.2,line width=13pt,rounded corners=1pt,color=asparagus}}

\usepackage[capitalize,noabbrev]{cleveref}
\usepackage{framed}

\theoremstyle{plain}
\newtheorem{theorem}{Theorem}[section]
\newtheorem{proposition}[theorem]{Proposition}
\newtheorem{lemma}[theorem]{Lemma}
\newtheorem{corollary}[theorem]{Corollary}
\theoremstyle{definition}

\newtheorem{assumption}[theorem]{Assumption}

\theoremstyle{remark}

\usepackage[textsize=tiny]{todonotes}

	\definecolor{comment}{HTML}{419d78} 
	\definecolor{asparagus}{HTML}{d33f49}
	\definecolor{blood}{HTML}{d33f49}
	\definecolor{lilly}{HTML}{d7c0d0}
	\definecolor{night}{HTML}{262730}

\icmltitlerunning{Meta-Learning Hypothesis Spaces}

\begin{document}

\twocolumn[
  \icmltitle{Meta-Learning Hypothesis Spaces for Sequential Decision-making}



\icmlsetsymbol{equal}{*}

\begin{icmlauthorlist}
\icmlauthor{Parnian Kassraie}{eth}
\icmlauthor{Jonas Rothfuss}{eth}
\icmlauthor{Andreas Krause}{eth}
\end{icmlauthorlist}

\icmlaffiliation{eth}{ETH Zurich, Switzerland}

\icmlcorrespondingauthor{Parnian Kassraie}{pkassraie@ethz.ch}

\icmlkeywords{Meta-Learning,  Confidence Bounds, Sequential Decision-making}

\vskip 0.3in
]



\printAffiliationsAndNotice{}  

\begin{abstract}
\looseness -1 Obtaining reliable, adaptive confidence sets for prediction functions (hypotheses) is a central challenge in sequential decision-making tasks, such as bandits and model-based reinforcement learning.   
These confidence sets typically rely on prior assumptions on the hypothesis space, e.g., the {\em known} kernel of a Reproducing Kernel Hilbert Space (RKHS). Hand-designing such kernels is error prone, and misspecification may lead to poor or unsafe performance. 
In this work, we propose to {\em meta-learn a kernel from offline data} (\mmkl). For the case where the unknown kernel is a combination of known base kernels, we develop an estimator based on structured sparsity. Under mild conditions, we guarantee that our estimated RKHS yields valid confidence sets that, with increasing amounts of offline data, become {\em as tight as those given the true unknown kernel}. We demonstrate our approach on the kernelized bandit problem (a.k.a.~Bayesian optimization), where we establish regret bounds competitive with those given the true kernel. We also empirically evaluate the effectiveness of our approach on a Bayesian optimization task.

\end{abstract}

\section{Introduction}

 \begin{figure}[t]
    \centering
    \includegraphics[width =\linewidth]{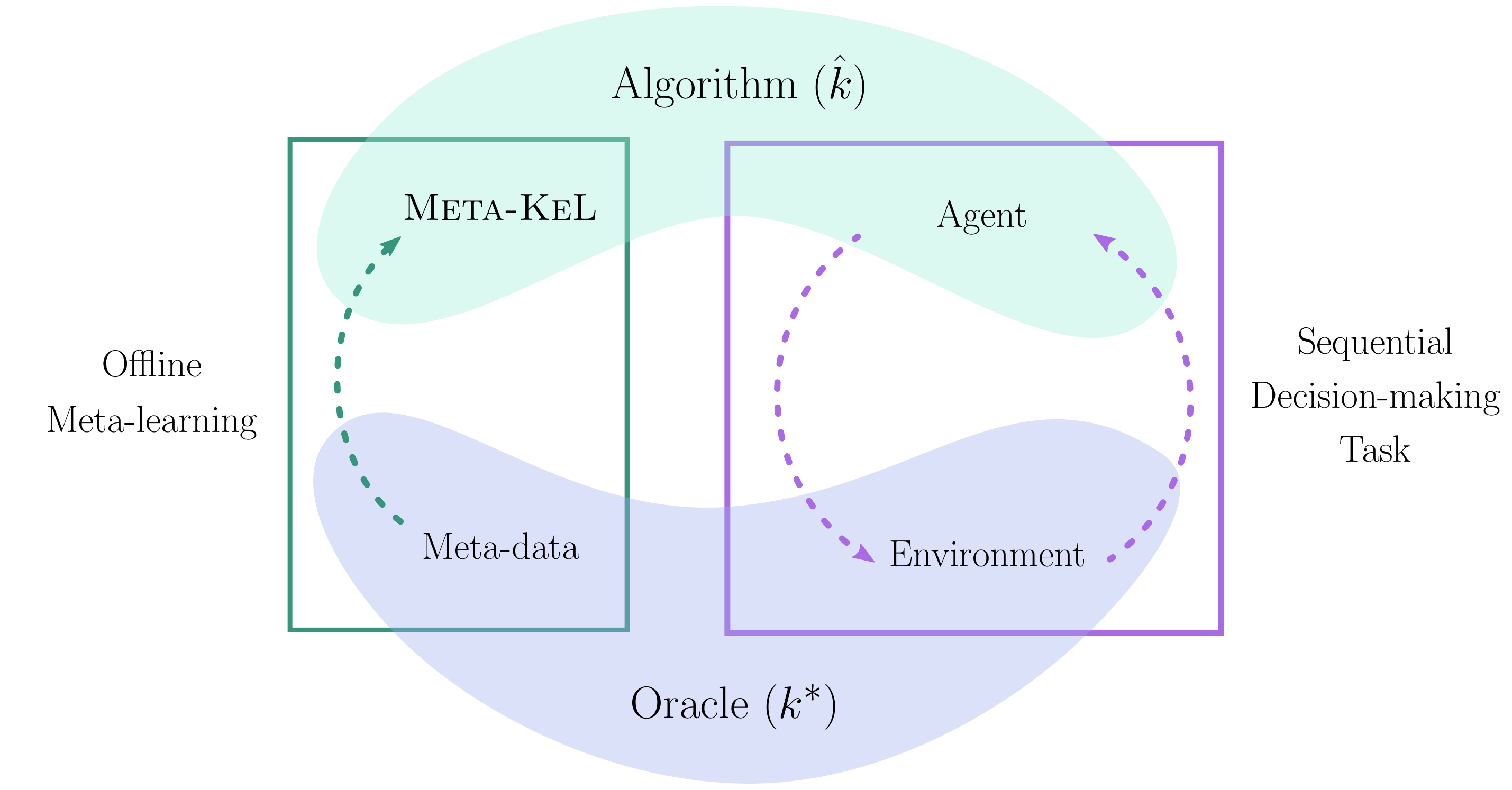}
    \caption{Overview of the described framework with $\tk$ as the true kernel function and $\khat$ as the solution to \mmkl. \vspace{-14pt}}
    \label{fig:mainidea}
\end{figure}

A number of well-studied machine learning problems such as bandits, Bayesian optimization (BO) and model-based reinforcement learning are characterized by an agent that sequentially interacts with an unknown, responsive system. Throughout the interaction, the agent's goal is to maximize the cumulative reward based on an unknown underlying function $f$.
Common to such sequential decision-making problems is an exploration-exploitation trade-off. That is, the agent needs to optimize its reward while, at the same time, learns more about the unknown function $f$. 
Confidence sets capture and quantify the uncertainty of the learner about $f$. 
Thus, they are an integral tool for directing exploration towards areas of high uncertainty and balancing it against exploitation. Moreover, in safety-critical applications, confidence sets are used to reason about the safety of actions. Thus, they are central to efficiency and safety of exploration.
In theoretical analysis of sequential decision-making algorithms, a common assumption is that $f$ resides in an RKHS with a \emph{known} kernel function. This assumption allows for the construction of the confidence sets.

\looseness -1 In practice, however, the true kernel is unknown and needs to be hand-crafted based on the problem instance. This is a delicate task, since the hand-crafted hypothesis space has to contain the unknown target function $f$. If this is not the case, the learner may be over-confident and converge to a sub-optimal policy, or incorrectly classify actions as safe. At the same time, we want the chosen hypothesis space to be as small so that the variance of the associated learner is low and the agent converges quickly. This constitutes a dilemma, where we need to trade off efficiency with a potential loss in consistency.

We approach this dilemma in a data-driven manner.
Many applications of sequential decision-making, such as hyper-parameter tuning with BO or online nonlinear control, are of repetitive nature. Often, there is available data from similar but not identical tasks which have been solved before.
Therefore, we propose to {\em meta-learn the kernel function, and thus the RKHS, from offline meta-data}. Our method, {\em Meta-Kernel Learning} (\mmkl), works with a generic (i.e., not necessarily i.i.d.) data model and may be applied to a variety of sequential decision-making tasks.

We formally analyze the problem when the true kernel is a combination of known base kernels.
We prove that the solution to \mmkl corresponds to an RKHS which contains the true function space (\cref{thm:RKHS_recovery}). Further, the meta-learned kernel has a sparse structure (\cref{prop:sparsity_bound}) which reduces the variance of the resulting learner, and makes the learner more efficient for solving the downstream sequential decision-making problem.
With mild assumptions on the data, we show that the meta-learned kernel yields any-time valid confidence bands matching the ones given oracle knowledge of the true kernel, as more samples of the meta-data are provided (\cref{thm:conf_int}). These provably reliable confidence estimates constitute the key contribution of our work and distinguishes \mmkl from prior attempts.

To demonstrate how \mmkl can be applied to a sequential task, we analyze a Bayesian optimization algorithm when it uses the meta-learned kernel and compare it to the same algorithm when it has knowledge of the true kernel, i.e., the oracle algorithm. By increasing size of the meta-learning data, the regret bound we obtain approaches the rate of the oracle (\cref{cor:regret_bound}).
\vspaceparagraph
\paragraph{Contributions} Our main contributions are:
\begin{itemize}
    \item We introduce \mmkl, a method for meta-learning the hypothesis space of a sequential decision task, which yields provably valid adaptive confidence sets.
     \item Our meta-learnt confidence bounds  converge to the ones estimated by the oracle at a $\gO(1/\sqrt{mn})$ rate. Here, $m$ is the number of tasks in the meta-data and $n$ is the number of samples per task. 
    \item Applied to BO,
    our results imply a sublinear regret guarantee for the \textsc{GP-UCB} algorithm using our meta-learned kernel.  
    This bound approaches that of the oracle algorithm as the amount of meta-data increases.
\end{itemize}

\section{Related Work}
\looseness -1 Numerous sequential decision-making methods rely on confidence sets for uncertainty quantification, e.g., UCB algorithms \citep{srinivas2009gaussian, chowdhury2017kernelized} for Bayesian optimization and bandits, safe exploration and various forms of RL \citep{berkenkamp2017safe, curi2020efficient, kakade2020information, sessa2020learning}. 
Most of these methods assume the true hypothesis space as given. However, in practice, we typically do not know the correct hypothesis space, e.g., in form of a kernel.
A body of recent work considers the unknown kernel setting and analyzes the effect of working with a misspecified hypothesis space \citep{wynne2021convergence, simchowitz2021bayesian, bogunovic2021misspecified}. Alternatively, \citet{wang2014theoretical} and \citet{berkenkamp2019no} propose to successively expand the hypothesis space throughout the course of BO so that the algorithm remains no-regret in a setting where the kernel lengthscale is unknown.


\looseness -1 Our work relates to meta-learning for Bayesian optimization. 
There is a recent line of algorithms that improve accuracy of base sequential learners via meta-learning, albeit without theoretical guarantees \citep{rothfuss2021pacoh,rothfuss2021meta}, or with mild guarantees in special cases \citep{kveton2020meta,boutilier2020differentiable}.
There are a number of results on updating bandit priors or policies by meta-learning, under problem settings different than ours. \citet{basu2021no} work with a sequence of multi-armed bandit tasks, and adaptively meta-learn the {\em mean} of the Gaussian prior used for the next task.
Others consider solving a number of structurally similar linear bandit tasks in parallel \citep{wang2017batched, cella2020meta, cella2021multi}. They propose how to efficiently update the policy when each learner has access to the data across all tasks. 
We give stronger guarantees in less restrictive setting, compared to \citet{wang2018regret} which analyzes the simple regret of the GP-UCB algorithm \citep{srinivas2009gaussian} for multi-armed and linear bandits, when the mean and variance of the Gaussian prior are unknown, and there is sufficient offline i.i.d.~data drawn from the same Gaussian distribution. 

\looseness -1 Our framework considers structural sparsity at the kernel level,  translating to group sparsity for the coefficients vectors, if applied to linear bandits. Thus our work relates to results on sparse linear bandits and Lasso bandits. In this area, \citet{bastani2020online} and \citet{hao2020high} give dimension-independent regret bounds for \emph{Explore-Then-Commit} algorithms under certain assumptions over the action set. This work does not consider offline data.


\looseness -1 We draw inspiration from the early Multiple Kernel Learning (MKL) literature, which focuses on kernel design for classification with SVMs \citep{bach2004multiple,gonen2011multiple,kloft2011lp,evgeniou2004regularized,cristianini2006kernel,ong2005learning}. 
In contrast, our key contribution is to derive adaptive confidence bounds from meta-learnt kernels for regression, even for non-i.i.d.~data.
Orthogonal to most prior works, we reduce the kernel learning problem to group Lasso and leverage the properties of the Lasso estimator, in particular seminal results of \citet{lounici2011oracle} and \citet{bach2008consistency}. Other relevant works on convergence properties of the group Lasso include \citet{koltchinskii2008sparse}, \citet{liu2009estimation} and \citet{bunea2013group}.

\section{Problem Statement}\label{sec:model}

\looseness -1 Consider a sequential decision-making problem, where the agent repeatedly interacts with an environment and makes observations 
\vspace{-3pt}
\begin{equation}\label{eq:BO_data_model}
\vy_t = f(\vx_t) + \varepsilon_t
\vspace{-2pt}
\end{equation}
of an unknown function $f: \gX \rightarrow \sR$  residing in an RKHS $\tH$ that corresponds to an {\em unknown} kernel function $\tk$.\footnote{Appendix \ref{app:rkhs} presents a compact refresher on the RKHS.}
We further assume that the function has a bounded kernel norm $\norm{f}_\tk \leq B$ and that the domain $\gX \subset \sR^{d_0}$ is compact.
\looseness -1 The observation noise $\varepsilon_t$ are i.i.d.~samples from a zero-mean sub-Gaussian distribution with variance proxy $\sigma^2$.
At every step $t$, the chosen input $\vx_t$ only depends on the history up to step $t$, denoted by the random sequence $H_{t-1} = \{ (\vx_\tau, y_\tau): \, 1 \leq \tau \leq t-1\}$. No further assumptions are made about the algorithm or the policy for choosing $\vx_t$.
Depending on the application, \cref{eq:BO_data_model} can serve different purposes: It can describe the stochastic reward model of a bandit problem, or it may be the transition dynamics of an RL agent in a stochastic environment.

\looseness -1 For solving such problems, a central prerequisite for numerous algorithms are {\em confidence sets} for $f(\vx)$ based on the history $H_{t-1}$ to balance exploration and exploitation at any step $t$.
 For any $\vx \in \gX$, the set $\gC_{t-1}(\vx)$ defines an interval to which $f(\vx)$ belongs with high probability such that,
 \[\sP\left(\forall\vx\in\gX:f(\vx) \in \gC_{t-1}(\vx)\right)\geq 1-\delta.\]
\looseness -1 The center of this interval reflects the current knowledge of the agent, relevant for exploitation, and the width corresponds to the uncertainty, guiding further exploration.
When the true kernel is known, an approach commonly used in the kernelized bandit literature \citep{abbasiyadkori2011, srinivas2009gaussian, russo2014learning} is to build sets of the form 
\begin{align}
    \gC_{t-1}(k;\vx) = [&\mu_{t-1}(k;\vx)-\nu_t\sigma_{t-1}(k;\vx), \label{eq:conf_set_def}\\ 
    &\mu_{t-1}(k;\vx)+\nu_t\sigma_{t-1}(k;\vx)]\notag
\end{align}
where the exploration coefficient $\nu_t$ depends on the desired confidence level $1-\delta$, and may be set based on the objective of the decision-making task. The functions $\mu_{t-1}$ and $\sigma_{t-1}$ set the center and width of the confidence set as
\begin{align}
    \mu_{t-1} (k;\vx)& = {\bm{k}}_{t-1}^T(\vx)({\bm K}_{t-1}+\bar\sigma^2\bm{I})^{-1}\vy_{t-1}  \label{eq:GPposteriors}\\
     \sigma^2_{t-1}(k; \vx) & =  k(\vx, \vx) - {\bm k}^T_{t-1}(\vx)({\bm K}_{t-1}+\bar\sigma^2\bm{I})^{-1}{\bm k}_{t-1}(\vx) \nonumber
\end{align}
where $\bar \sigma$ is a constant, $\vy_{t-1} = [y_\tau]_{\tau < t}$ is the vector of observed values, $\bm{k}_{t-1}(\vx) = [ k(\vx, \vx_\tau)]_{\tau < t}$, and ${\bm K}_{t-1} = [ k(\vx_i, \vx_j)]_{i,j < t}$ is the kernel matrix. 
Hence working with the right kernel function plays an integral role in constructing well-specified sets. Since, in practice, the true kernel $\tk$ is not known by the learner, most approaches use a hand-designed kernel that suits the problem instance at hand or conservatively pick an expressive kernel that constructs a rich RKHS which is very likely to contain $f$. There are a number of empirical approaches for selecting the kernel, e.g., cross-validation or maximizing the marginal likelihood. However, such methods tend to be unreliable when the available data is non-i.i.d.~and comes from sequential learning tasks.

Addressing the issue of selecting a correct and yet efficient kernel, we pursue a data-driven approach and {\em meta-learn a kernel that provably yields valid confidence intervals}. This guarantee is valid regardless of how the meta-data is gathered, as long as it satisfies some basic conditions discussed later in Assumptions \ref{cond:betamin}~and~\ref{cond:RE}.
 We consider an offline collection of datasets $\gD_{n,m}= \{(\vx_{s,i},y_{s,i})_{i\leq n}\}_{s \leq m}$ from $m$ possibly non-i.i.d. tasks, each with a sample size $n$.
Suppose, for each task $s$, labels are generated by
\begin{equation}\label{eq:meta_data_model}
    y_{s,i} = f_{s}(\vx_{s,i}) + \varepsilon_{s,i}
\end{equation}
for $i \leq n$, where $\varepsilon_{s,i}$ are zero-mean i.i.d.~sub-Gaussian noise with variance proxy $\sigma^2$.
We assume the tasks are related by the fact that all $f_s: \gX \rightarrow \sR$ come from the same function class $\tH$ and have a bounded RKHS norm $\norm{f_s}_\tk \leq B$.
We do not make any assumptions on the policy based on which the points $\vx_{s,i}$ are chosen.
\vspaceparagraph
\paragraph{Assumptions}
Our analysis requires some assumptions on the kernel function. In particular, we assume that $\tk$ is a finite combination of known base kernels,
\begin{equation}\label{eq:kernel_dec}
    \tk(\vx,\vx') = \sum_{j =1}^p \teta_j k_j(\vx, \vx'),
\end{equation}
where the weight vector $\tveta \geq 0$ is {\em unknown}. 
Without loss of generality,
 we assume that $\tk$ and the base kernels are all {\em normalized}, i.e., $\norm{\tveta}_1\leq 1 $ and $k_j(\vx,\vx') \leq 1$ for all $1\leq j \leq p$ and $\vx ,\, \vx' \in \gX$. The weight vector $\tveta$ is potentially {\em sparse}, since not all the candidate kernels $k_j$ actively contribute to the construction of $\tk$.
We use $J_\tk = \{ 1\leq j\leq p: \teta_j \neq 0\}$ to refer to the group of base kernels that are present in $\tk$. 
The sparse construction of $\tk$ imposes favorable structure on the meta-data, which essentially allows us to meta-model-select the hypothesis space and recover the true sparsity pattern denoted by $J_\tk$.
We further assume that each $k_j$ has a $d_j$-dimensional feature map, i.e., $k_j(\vx,\vx') = \vphi_j^T(\vx)\vphi_j(\vx')$, where $\vphi_j \in \sR^{d_j}$. For the scope of this paper, we assume that $\dmax < \infty$, where $\dmax := \max_{j\leq p} d_j$.
In this finite regime, the analysis can also be carried out in a finite-dimensional vector space. Nevertheless, we use a function space notation since, even though our theory focuses on the finite-dimensional setting, empirically our approach is also applicable to kernels with infinite dimensional feature map.\footnote{Meta-learning the hypothesis space in the $p \rightarrow \infty$ limit will be challenging. However, we expect to obtain an extension to infinite dimensional base-kernels in future work.\looseness-1} 

Let $\vphi(\vx)$ denote the $d$-dimensional feature map for $\tk$ where $d =  \sum_{j=1}^pd_j$ and
\vspace{-3pt}
\[\vphi(\vx) = \left(\sqrt{\teta_1}\vphi_1^T(\vx),\cdots, \sqrt{\teta_p}\vphi_p^T(\vx)\right)^T.\]
For each task $s$, the function $f_s$ is contained in $\tH$. By Mercer's theorem $f_s$ may be decomposed as
\begin{equation}\label{eq:f_decompose}
f_s(\vx) =  \vphi^T(\vx) \tvbeta_s  = \sum_{j = 1} ^ p \sqrt{\teta_j} \vphi_j^T(\vx) \tvbeta_s\gj,
\end{equation}
where $\tvbeta_s \in \sR^d$ is the coefficients vector of task $s$ and $\tvbeta_s\gj \in \sR^{d_j}$ is the sub-vector corresponding to kernel $k_j$.
It is not possible to meta-select a base kernel $k_j$ which has not contributed to the generation of the meta-data. Therefore, if a base kernel is active in the construction of $\tH$, it is only natural to assume that there is some task in the meta-data which reflects this presence. 
More formally, we assume that, for any $j \in J_\tk$, there exists some $s \leq m$ where $\tvbeta_s\gj \neq 0$. 
We define $\tvbeta = (\tvbeta_1{^T},\cdots,\tvbeta_m{^T})^T \in \sR^{md}$ as the concatenated coefficients vector for all tasks. 
To refer to the group of coefficients that correspond to kernel $k_j$ across all tasks, we use $\tvbeta\gj =  ((\tvbeta_1\gj)^T,\cdots,(\tvbeta_m\gj)^T)^T\in \sR^{md_j}$. 
 \cref{tab:notations} presents a compact guide to the notation introduced here. 
Our next assumption guarantees that the meta-learning problem is not ill-posed.
\begin{assumption}[Group Beta-min Condition]\label{cond:betamin}
There exists $c_1>0$ s.t. for all $j\in J_\tk$ it holds that $\norm{\tvbeta\gj}_2 \geq c_1$.
\end{assumption}
\looseness -1 
This assumption is inevitable for recovering the sparsity pattern from empirical data and it is widely used in the high-dimensional statistics literature \citep[e.g.,][]{buhlmann2011statistics, zhao2006model, van2011adaptive}. \cref{cond:betamin} implies that for $j$ to be in $J_\tk$, the coefficients vector corresponding to kernel $k_j$ can not be zero or arbitrarily close to zero. In practice, $\norm{\tvbeta\gj}_2$ has to be comparable with the noise level for the activity of a base kernel not to be mistaken with randomness.

\section{Meta-learning the Hypothesis Space (\mmkl)}\label{sec:meta}
In the following section, we present our formulation of the meta-learning problem and analyze the properties of the learned hypothesis space. We meta-learn the kernel by solving the following optimization problem. Then, we set the hypothesis space of the downstream learning algorithm to be the RKHS of the meta-learned kernel.
\begin{equation}\label{eq:nonparam_opt}
\begin{split}
    \min_{\veta,f_1,\dots,f_m} & \,\frac{1}{m}\sum_{s=1}^m \left[\frac{1}{n}\sum_{i=1}^n \left(y_{s,i} - f_{s}(\vx_{s,i})\right)^2\right]
    \\
    &\quad\quad+ \frac{\lambda}{2} \sum_{s = 1}^m \norm{f_s}^2_k + \frac{\lambda}{2}\norm{\veta}_1\\
     \text{s.t. } & \,\forall s: f_s \in \gH_k,\,  k = \sum_{j =1}^p \eta_j k_j, 0 \leq \veta
\end{split}
\end{equation}
We will refer to this problem as \emph{Meta-Kernel Learning} (\mmkl). The first part of the objective is similar to the kernel ridge regression loss, and accounts for how well a series of regularized $f_s$ fit the meta-data. The last term regularizes our choice of the kernel function. We use $\ell_1$-norm regularization for $\veta$ to implicitly perform meta-model-selection. As shown in~\cref{prop:sparsity_bound}, the meta-learned kernel will reflect the sparsity pattern of the true kernel.
The optimization problem (\ref{eq:nonparam_opt}) is convex and admits an efficient solution, as explained next.

We first introduce a vectorized formulation of \cref{eq:nonparam_opt}. 
Let $\vy_s \in \sR^{n}$ denote the observed values for a task $s$ and $\vy = (\vy_1^T,\cdots,\vy_m^T)^T \in \sR^{mn}$ the multi-task stacked vector of observations. 
We then design a multi-task feature matrix. 
We define $\mPhi$ to be a $mn\times md$ block-diagonal matrix, where each block $s$ corresponds to $\mPhi_s = (\vphi(\vx_{s,1}), \cdots, \vphi(\vx_{s,n}))^T$, the $n\times d$ feature matrix of task~$s$. Figure \ref{fig:glasso_formulation} provides an illustration thereof. As shown in Proposition~\ref{prop:eta_hat}, this vectorized design brings forth a parametric equivalent of \mmkl, which happens to be the well-known Group Lasso problem.
\begin{proposition}[Solution of \mmkl]\label{prop:eta_hat}
Let $k=\sum_j\etahat_jk_j$ be a solution to Problem (\ref{eq:nonparam_opt}). Then, for all $1\leq j \leq p$, it holds that
\[
\etahat_j =  \norm{\vbetahat\gj}_2
\]
with $\vbetahat = (\vbetahat\gj)_{j\leq p}$ as the solution of the following convex optimization problem:
\begin{equation}\label{eq:glasso_opt_multi}
\min_{\vbeta} \frac{1}{mn} \norm{\vy-\mPhi\vbeta}_2^2 + \lambda \sum_{j=1}^p \norm{\vbeta\gj}_2.
\end{equation}
\end{proposition}
We show this equivalence by eliminating $\veta$. We use a trick introduced by \citet{bach2004multiple}, which, for $w, v \in \sR$ states 
$
2\abs{w} = \min_{v\geq 0} w^2/v + v.
$
The proof is given in Appendix \ref{app:eta_trick_proof}. Problem (\ref{eq:glasso_opt_multi}) can be optimized by any Group Lasso solver.  \citet{bach2011optimization} present a number of coordinate descent algorithms which efficiently find the solution.

Before introducing the meta-learned kernel $\khat$, we note that Reproducing Kernel Hilbert Spaces are equivalent up to scaling of the kernel function. For $c>0$, both $\gH_k$ and the scaled version $\gH_{ck}$ contain the same set of functions. Going from $\gH_k$ to $\gH_{ck}$, the RKHS norm of any member $f$ would scale by $1/c$, i.e. $\norm{f}_k = c\norm{f}_{ck}$. Hence, the norm $\norm{\vetahat}_1$ will be irrelevant when meta-learning the function space. This norm can be scaled or normalized, and still yield the same hypothesis space, only with a scaled operator norm.
For consistency of notation, we define $\khat$ as follows. For any two points $\vx,\,\vx'\in \gX$, set
\begin{equation}\label{eq:meta_kernel}
    \khat(\vx,\vx') = \sum_{j=1}^p\frac{\etahat_j}{c_1}\vphi_j^T(\vx)\vphi_j(\vx'),
\end{equation}
where $c_1$ is the same constant as in \cref{cond:betamin}.
We emphasize that this scaling does not impose a new assumption on the problem as it is done only to simplify theorem statements. 
We denote the set of base kernels active in $\khat$ with $J_\khat = \{ 1\leq j\leq p: \etahat_j \neq 0\}$. 
The meta-learned hypothesis space will then be $\Hhat$, the RKHS which corresponds to $\khat$. 

\vspaceparagraph
\paragraph{Properties of the Meta-learned Hypothesis Space}
The meta-learned $\khat$ can be used as the kernel function for a model-based sequential decision making algorithm, as illustrated in \cref{fig:mainidea}. Our goal is to analyze how this choice of kernel affects the success of the algorithm, compared to the oracle algorithm with access to the unknown kernel. To this end, we discuss properties of $\khat$. 

For our main result, we require a final technical assumption to ensure that our meta-data is sufficiently diverse.
Consider some vector $\vb \in \sR^{md}$ that adheres to the same group structure as $\tvbeta$. For a set of group indices $J \subset \{1,\cdots,p\}$,  we define $\vb_J := (\vb\gj)_{j \in J}$ as the sub-vector indicated by the groups in $J$. 

\begin{assumption}[Relaxed Sufficient Exploration] \label{cond:RE}
There exists $\kappa = \kappa(s)>0$ that satisfies
\begin{align*}
    \kappa & \leq \min_{J, \vb} \frac{\norm{\Phi\vb}_2}{\sqrt{mn}\norm{\vb_J}_2},\\
     \text{s.t. }&\sum_{j \notin J} \norm{\vb\gj}_2\leq 3 \sum_{j \in J}\norm{\vb\gj}_2,\,\vb\neq 0,\,\abs{J}\leq s.
\end{align*}
\end{assumption}
This condition makes sure that the meta-training data is not degenerate, e.g., no two points $\vx_{s,i}$ and $\vx_{s,j}$ in the meta-data are identical or too close. 
Assumption~\ref{cond:RE} is immediately fulfilled if the minimum eigenvalue of $\mPhi$ is positive, i.e., if there exists some $\kappa>0$ where $\lambda_{\mathrm{min}}(\mPhi)\geq \kappa$. This stronger version is sometimes referred to as Sufficient Exploration \citep{basu2021no, zhou2020neural}. 
In the Lasso literature, \cref{cond:RE} is commonly referred to as Restricted Eigenvalue Assumption \citep{bickel2009simultaneous,buhlmann2011statistics,javanmard2014confidence}. It is also often required to hold for the kernel matrix in the sparse linear bandits literature \citep{bastani2020online,wang2018minimax,hao2020high, kim2019doubly}. Lastly, \cref{cond:RE} is satisfied with high probability if $\mPhi$ is an i.i.d.~sub-Gaussian random matrix, since the eigenvalues of such random matrices are bounded away from zero  \citep[Theorem 4.6.1]{vershynin2018high}.

\begin{theorem}[Hypothesis Space Recovery]\label{thm:RKHS_recovery}
Set $0\leq \delta \leq 1$, and choose $\lambda$ such that,
\[
\lambda \geq \frac{4\sigma}{\sqrt{mn}}\sqrt{1 + \frac{2}{m}\left( \log (2p/\delta) + \sqrt{m\dmax \log( 2p/\delta)}\right)}.
\]
If $\vert J_\tk \vert \leq s$ and Assumption \ref{cond:RE} holds with $\kappa(s)$, then $\tH \subseteq \Hhat$ and  
\[
\norm{f}_\khat \leq \norm{f}_\tk \Big(1 + \epsilon(n,m) + o\big(\epsilon\left(n,m\right)\big)\Big),
\]
with probability greater than $1-\delta$, if $n$ and $m$ are large enough to satisfy $\epsilon(n,m) \leq c_1$. The absolute constant $c_1$ is defined in \cref{cond:betamin} and
\begin{align*}
\epsilon(n,m) := & \frac{32\sigma s}{\kappa^2(s)\sqrt{mn}}\\
\times &\sqrt{1 + \frac{2}{m}\left( \log (2p/\delta) + \sqrt{m\dmax \log( 2p/\delta)}\right)}.
\end{align*}
\end{theorem}
The proof is given in \cref{app:proof_rkhs}.
Theorem \ref{thm:RKHS_recovery} states that, provided enough meta-data, $\tH$ is contained in $\Hhat$ with high probability and $\delta$, the probability of failure in recovery, decreases as $m$ and $n$ grow (See \cref{proof:A_happens}).
In addition the RKHS norm of any $f \in \tH$, can be bounded arbitrary well by $\norm{f}_\khat$, since the error $\epsilon(n,m)$ decreases at a $\gO(s/\sqrt{mn}\sqrt{1+m^{-1}\log p})$ rate.
This matches the tightest rate for coordinate-wise convergence of the Group Lasso estimator under the same set of assumptions \citep{lounici2011oracle, bunea2013group}. The theorem implies that the meta-learner benefits more from increasing the number $m$ of meta-data tasks, rather than increasing the sample size $n$ of each task, since $\epsilon(n,m)$ shrinks faster with $m$ compared to $n$.
Note that increasing either $m$ or $n$ will result in convergence and therefore this theorem also holds for the classic offline kernel learning setup when the dataset consists of a single learning task ($m=1$). 


\vspaceparagraph
\paragraph{The Benefit of Structural Sparsity} 
Consider a conservative hand-picked kernel function 
\begin{equation}\label{eq:k_full}
    \kfull = 1/p\sum_{j=1}^p\vphi^T_j(\vx)\vphi_j(\vx'),
\end{equation} 
which does not use any meta-data and instead incorporates all the considered base kernels. 
When $p$ and $\dmax$ are finite, $\tH$ is contained in $\gH_\kfull$ and the hand-picked hypothesis space is not misspecified. 
However, working with an overly large hypothesis space has downsides. 
Consider using $\kfull$ to estimate a function $f \in \tH$.
Then every base kernel, including $k_j$ with $j \notin J_\tk$, appears in the construction of the estimator. 
These terms contribute to the estimation error and increase the variance of the function estimate. This slows down the rate of convergence, compared to the case where only active $k_j$ are present in the kernel function.
By meta-learning $\hat{k}$ via \mmkl, we can eliminate irrelevant candidate kernels and produce a structurally sparse hypothesis space. \cref{prop:sparsity_bound} guarantees this property. 
Its proof is given in \cref{proof:sparsity_bound}.
\begin{proposition}[Bound on structural sparsity of $\khat$]\label{prop:sparsity_bound}
Set $0<\delta<1$ and choose $\lambda$ according to Theorem \ref{thm:RKHS_recovery}.
Let $\vert J_\tk\vert \leq s$ be the number candidate kernels that contribute to $k^*$. If Assumption \ref{cond:RE} holds with $\kappa(s)$, then with probability greater than $1-\delta$, the number of kernels active in $\hat{k}$ is bounded by
\[
\vert J_\khat \vert \leq \frac{64s}{mn\kappa^2(s)}
\]
which implies that if $mn > \frac{64s}{p\kappa^2(s)}$, then with the same probability
  \[\Hhat \subsetneq \gH_\kfull.\]
\end{proposition}

\begin{figure}[ht]
    \centering
    \includegraphics[width =0.7\linewidth]{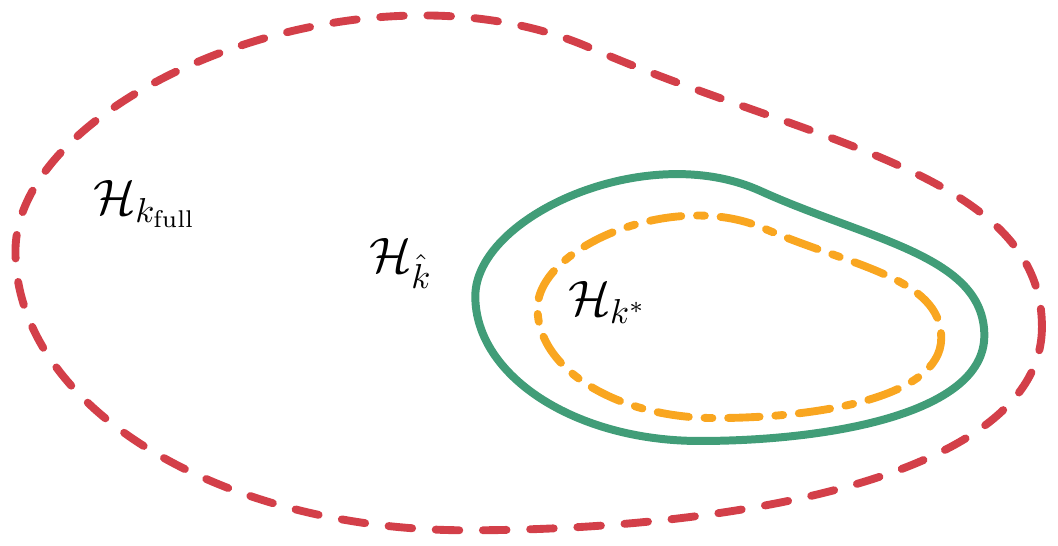}
    \caption{The oracle $\tH$ (Eq.~\ref{eq:kernel_dec}), the meta-learned $\Hhat$ (Eq.~\ref{eq:meta_kernel}) and the hand-picked $\gH_\kfull$ (Eq.~\ref{eq:k_full}) hypothesis spaces (informal)}
    \label{fig:hypotheses}
\end{figure}

Hence, in the presence of enough meta-data, $\Hhat$ is a strict subset of $\gH_\kfull$, and therefore
\[
\tH \stackrel{\mathrm{w.h.p.}}{\subseteq} \Hhat \stackrel{\mathrm{w.h.p.}}{\subsetneq} \gH_\kfull
\]
where the left relation is due to Theorem \ref{thm:RKHS_recovery}. Figure \ref{fig:hypotheses} illustrates the nested sets. We conclude that our meta-learned hypothesis space has favorable properties: it contains the true hypothesis space, and it is sparse in structure, in particular, smaller than the conservative candidate space.

The fact that $\Hhat$ is smaller than $\gH_\kfull$ reduces the complexity of the downstream learning problem and yields faster convergence rates. We provide an example of this effect in \cref{sec:bandit}, where we analyze a Bayesian optimization problem, and establish how choosing $\khat$ improves upon $\kfull$. Finally, our experiments (e.g. \cref{fig:regret_bo}) support the claim that in practice the BO algorithm is faster in finding the optimum when it uses the meta-learned kernel.

\section{Sequential Decision-making with \mmkl}\label{sec:BO}

We now analyze the effect of using $\khat$ as kernel function in the downstream sequential decision-making problem. 
We adopt the common construction of confidence sets given in \cref{eq:conf_set_def}, and define $\hat\gC_{t-1}(\vx) :=\gC_{t-1}(\khat; \vx)$. We let $\hat\mu_{t-1}(\vx) := \mu_{t-1}(\khat; \vx)$, and $\hat\sigma_{t-1}(\vx) := \sigma_{t-1}(\khat; \vx)$, where $\mu_{t-1}(k;\vx)$ and $\sigma_{t-1}(k; \vx)$ are as defined in \cref{eq:GPposteriors}, with time-varying $\bar \sigma^2 = 1 + 2/t$.\footnote{\looseness -1 The functions $\hat\mu_{t-1}$ and $\hat\sigma_{t-1}$ are the posterior mean and variance of $\mathrm{GP}(0, \khat)$, conditioned on $H_{t-1}$, with noise variance $\bar\sigma^2$.}

Theorem \ref{thm:conf_int} shows that for the right choice of $\nu_t$, the set $\hat C_{t-1}(\vx)$ is a valid confidence bound for any $f \in \tH$, evaluated at any $\vx\in \gX$, at any step $t$, with high probability.
\begin{theorem}[Any-time Valid Confidence Bounds with \mmkl]\label{thm:conf_int}
Let $f \in \tH$ with $\norm{f}_\tk \leq B$, where $\tk$ is unknown. Under the assumptions of Theorem \ref{thm:RKHS_recovery}, with probability greater than $1-\delta$, for all $\vx \in \gX$ and all $t \geq 1$,
\begin{equation*}
\begin{split}
        \vert \hat\mu_{t-1}(\vx) - &f(\vx)\vert \leq  \hat\sigma_{t-1}(\vx) \Bigg( B\left(1 + \frac{\epsilon(n,m)}{2c_1} \right)\\
        &+ \sigma\sqrt{\hat d\log\left(1 +  \frac{\bar\sigma^{-2}t}{c_1} \right) +2+2\log(1/\delta)}\Bigg)
\end{split}
\end{equation*}
where $\hat d= \sum_{j\in J_\khat}d_j$ and $\bar \sigma^2 = 1+ 2/t$.
\end{theorem}
The proof is given in Appendix \ref{app:BO}. As discussed in Section~\ref{sec:meta}, the $\epsilon(n,m)/2c_1$ term shrinks faster than $\gO(1/\sqrt{mn})$ and $\hat d$ approaches $d^* = \sum_{j\in J_\tk}d_j$ at a similar rate. Therefore, Theorem \ref{thm:conf_int} presents a tight confidence bound relative to the case when $\tk$ is known by the agent. In this case, due to \citet{chowdhury2017kernelized}, Theorem 2, the $1-\delta$ confidence bound would be,
\begin{equation*}
    \begin{split}
\vert \mu_{t-1}(\vx) - & f(\vx)\vert \leq \sigma_{t-1}(\vx)\Big( B \,+ \\
& \sigma\sqrt{d^*\log\big(  1 +  \bar\sigma^{-2}t\big) +2+2\log(1/\delta))}\Big).
    \end{split}
\end{equation*}
where the mean and variance functions are defined by $\mu_{t-1}(\vx) := \mu_{t-1}(\tk; \vx)$ and $\sigma_{t-1}(\vx) := \sigma_{t-1}(\tk; \vx)$ with $\bar\sigma^2 = 1+ 2/t$.
We conclude that the base learner does not require knowledge of the true kernel for constructing confidence sets, as long as there is sufficient meta-data available. \cref{thm:RKHS_recovery} quantifies this notion of sufficiency.

\vspaceparagraph
\paragraph{Case Study: Bayesian Optimization}\label{sec:bandit}

As an example application, we consider the classic Bayesian optimization problem, but in the case  where $\tH$ is unknown. This example illustrates how \cref{thm:conf_int} may be used to prove guarantees for a decision-making algorithm, which uses the meta-learned kernel due to a lack of knowledge of $\tk$.
We follow the setup and BO notation of \citet{srinivas2009gaussian}. 
The agent seeks to maximize an unknown reward function $f$, sequentially accessed
as described in \cref{eq:BO_data_model}.
Their goal is to choose actions $\vx_t$ which maximize the cumulative reward achieved over $T$ time steps. This is equivalent to minimizing the cumulative regret $R_T = \sum_{t=1}^T [f(\vx^*) - f(\vx_t)]$, where $\vx^*$ is a global maximum of $f$. Note that if $R_T/T \rightarrow 0$ as $T \rightarrow \infty$ then $\max_{1\leq t\leq T} f(\vx_t)\to f(\vx^*)$, i.e., the learner converges to the optimal value. We will refer to this property as \emph{sublinearity} of the regret.
In the spirit of the \textsc{GP-UCB} algorithm \citep{srinivas2009gaussian}, 
we choose the next point by maximizing the upper confidence bound as determined by Theorem~\ref{thm:conf_int}
\begin{equation}\label{eq:gpucb_policy}
    \vx_t = \argmax_{\vx \in \gX} \hat\mu_{t-1}(\vx) + \nu_t\hat\sigma_{t-1}(\vx)
\end{equation}
where a suitable choice for $\nu_t$ is suggested in \cref{cor:regret_bound}. 

\begin{corollary}[A Regret Bound with \mmkl]\label{cor:regret_bound} Let $\delta \in (0,1)$. Suppose $f\in \tH$ with $\norm{f}_\tk \leq B$ and that values of $f$ are observed with zero-mean sub-Gaussian noise of variance proxy $\sigma^2$. Then, with probability greater than $1-\delta$, \textsc{GP-UCB} used together with $\khat$ satisfies
\begin{align*}
 R_T= \gO\Bigg( \sqrt{\hat d T\log T}  &\Big( B\big(1 + \epsilon(n,m)\big) \\
 & + \sqrt{\hat d \log T+ \log1/\delta}\Big)\Bigg) 
\end{align*}
 provided that the exploration coefficient is set to
\begin{align*}\label{eq:gp_ucbcoef}
    \nu_t = & B\big(1 + \epsilon(n,m)/2c_1 \big) \\
    & + \sigma\sqrt{\hat d\log\left(1 +  \bar\sigma^{-2}t/c_1 \right) +2+2\log(1/\delta)}.
\end{align*}
\end{corollary}
The proof is straightforward. Conditioned on the event that $f \in \Hhat$, we may directly use the regret bound of \citet{chowdhury2017kernelized}. Then, by \cref{thm:RKHS_recovery}, we calculate the probability of this event (\cref{proof:regret_bound}).
The Corollary relies on knowledge of a bound $B$ on $\norm{f}_\tk$. However, using techniques of \citet{berkenkamp2019no} it is possible to adapt it even to {\em unknown} $B$ at increased (but still sublinear) regret.

\cref{cor:regret_bound} shows that \textsc{GP-UCB} using the meta-learned kernel guarantees sublinear regret. We obtain a $\gO(\hat dB\log T\sqrt{T})$ rate for the regret which is tight compared to the $\gO(d^*B\log T\sqrt{T})$ rate satisfied by the oracle.
It is insightful to compare this convergence result to a scenario where the hypothesis space is misspecified. For a reward function $f \notin \Hhat$, \citet{bogunovic2021misspecified} show that the learner will not converge to the global optimum, since the cumulative regret has a lower bound of linear order $\gO(T\sqrt{\log T})$. 
Corollary \ref{cor:regret_bound} suggests that by using a sparse kernel we can potentially find the optimal policy faster compared to when the complex kernel $\kfull$ is used. Recall that $d = \sum_{j=1}^pd_j$, by Theorem 2 of  \citet{chowdhury2017kernelized} the regret of \textsc{GP-UCB} used together with $\kfull$ is bounded by $\gO(dpB\log T\sqrt{T})$, since $\norm{f}_\kfull = p\norm{f}_\tk$. Therefore, using the meta-learned kernel improves the regret bound by a factor of $\hat d/(dp)$, implying that the solution may be found faster. The results of our experiments in
\cref{fig:regret_bo} support this argument. 

Note that our approach to guarantee a sublinear regret for \textsc{GP-UCB} without oracle knowledge of $\tk$ naturally generalizes to other sequential decision tasks. In particular, any theoretical result relying on 
RKHS confidence intervals with a {\em known} kernel can be immediately extended to use those of the meta-learned kernel.



\section{Experiments} \label{sec:experiments}
In this section, we provide experiments to quantitatively illustrate our theoretical contribution.  

\begin{figure}[t]
    \centering
    \includegraphics[width =0.95 \linewidth]{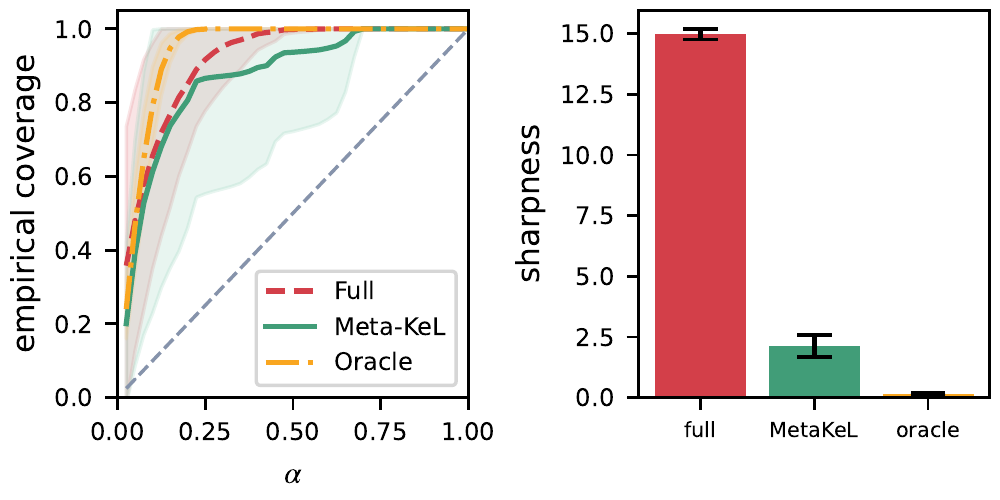}
    \vspaceexpfigure
    \caption{Calibration (left) and sharpness (right) experiment for confidence sets given 4 training samples. Averaged over $50$ runs, $\khat$ always gives tight valid confidence intervals.}
    \label{fig:calibrated_set}
\end{figure}
\begin{figure}[t]
    \centering
    \includegraphics[width =0.85 \linewidth]{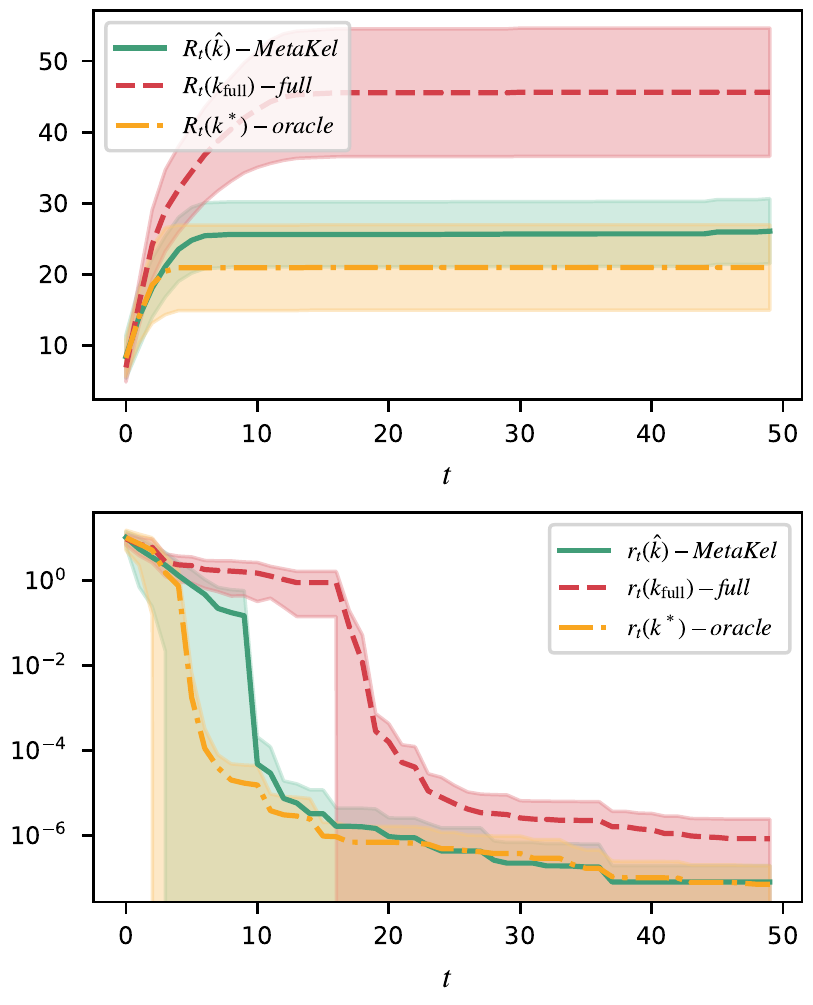}
    \vspaceexpfigure
    \caption{ Simple and cumulative regret for \textsc{GP-UCB}. 
    The algorithm converges at a slower pace when using $\kfull$.}
    \label{fig:regret_bo}
\end{figure}
\begin{figure}[t]
    \centering
    \includegraphics[width =0.9 \linewidth]{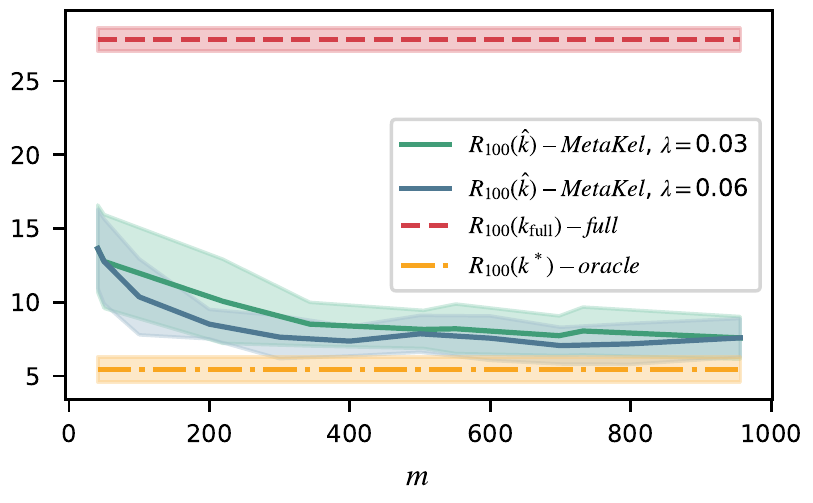}
    \vspaceexpfigure
    \caption{ The regret of \textsc{GP-UCB} used with $\khat$ approaches the oracle regret as the number of offline tasks increase.
    }
    \label{fig:bo_consistency}
\end{figure}

\vspaceparagraph
\paragraph{Experiment Setup (1D)} We create a synthetic dataset based on our data model, Equations~(\ref{eq:BO_data_model})~and~(\ref{eq:meta_data_model}). We first limit the domain to the $1$-dimensional $\gX= [-1,1]$ and use Legendre polynomials $P_j$ as our features $\vphi_j$.
The sequence $(P_j)_{j \geq 0}$ is a natural choice, since it provides an orthonormal basis for $L^2(\mathcal{X})$. Moreover, Legendre polynomials are eigenfunctions to dot-product kernels such as the Neural Tangent Kernel \citep{jacot2018neural}. 
We let $\tk(x, x') = \sum_{j\in J_\tk}\teta_j P_j(x)P_j(x')$, where $J_\tk$ is a random subset of $\{1,\cdots, p\}$. Each $\teta_j$ is sampled independently from the standard uniform distribution and the vector $\teta$ is then normalized. Across all experiments, we set $p=20$ and $s = \vert J_\tk\vert =5$.
To sample the meta-data $\gD_{n,m}$, we choose $m$ independent random subsets of $J_\tk$ and generate the functions $f_s$ via \cref{eq:f_decompose} where $\tvbeta\gj$ are drawn from an i.i.d.~standard uniform distribution. We then scale the norm $\norm{f}_\tk$ to $B = 10$. 
The data for a single task, i.e., $(x_{s,i}, y_{s,i})_{i \leq n}$, is then created by uniformly drawing ~i.i.d.~samples from the domain $\gX$ and evaluating $f_s$ at those points. 
We add Gaussian noise with standard deviation $\sigma = 0.01$ to all data points.
\cref{fig:functions} in the appendix shows how random $f_s$ may look like. For all experiments we set $n=m=50$ unless stated otherwise.
To meta-learn $\khat$, we solve the vectorized \mmkl problem (Eq. \ref{eq:glasso_opt_multi}) over $\vbeta\gj$ with \textsc{Celer}, a fast solver for the group Lasso \citep{celer2018}, and then set $\vetahat$ according to Proposition \ref{prop:eta_hat}. We set $\lambda = 0.03$, such that it satisfies the condition of \cref{thm:RKHS_recovery}. As shown in \cref{fig:lambda_choice} in the appendix, the choice of $\lambda$ has little effect on the performance of the algorithm.

\vspaceparagraph
\paragraph{Confidence Set Experiment} \looseness -1 We perform calibration and sharpness experiments to assess the meta-learned confidence sets \citep{gneiting2007probabilistic}. \cref{fig:calibrated_set} presents the result.
To obtain an $\alpha$-confidence interval for some $f(x)$ using a kernel $k$, we assume a $f\sim\mathrm{GP}(0, k)$ prior and calculate the $\alpha$-quantile of the posterior after observing $4$ noisy i.i.d. draws from the function. 
For each hypothesis, the $y$-axis of the left plot shows the empirical coverage of the confidence sets, i.e., the fraction of test points contained in the $\alpha$-confidence intervals for varying levels $\alpha$. In this plot, if a curve were to fall below the $x=y$ line, it would have implied insufficient coverage and hence over-confident sets.
The plot on the right shows the posterior variance averaged across all test points. This quantity, referred to as \emph{sharpness}, reflects the width of the confidence bands. 
Figure \ref{fig:calibrated_set} demonstrates that the meta-learned confidence sets are well-calibrated for the entire range of confidence-levels and are tight relative to the true sets. In contrast, $\kfull$ yields conservative confidence sets, due to considering polynomials $P_j$ that do not contribute to the construction of $f(x)$.
We use $1000$ test points for calculating the empirical averages.
The plot shows the values averaged over $50$ runs, where for each the kernel $\tk$ and the data are generated from scratch. 

\vspaceparagraph
\paragraph{Regret Experiment}We verify the performance of \textsc{GP-UCB} when used together with $\khat$. We generate the random reward function $f$ in a manner similar to $f_s$ of the meta-data. The BO problem is simulated according to \cref{eq:BO_data_model}, and the actions are selected via \cref{eq:gpucb_policy}. 
\cref{fig:sample_bo} in the appendix shows how this algorithm samples the domain and how the confidence estimates shrink by observing new samples.
Keeping the underlying random $\tk$ fixed, we generate $100$ random instances of the meta-learning and the BO problem. In \cref{fig:regret_bo} we present the average regret and its standard deviation. In these plots, the simple regret of \textsc{GP-UCB} with a kernel $k$ is labeled $r_t(k) = f(x^*)-\max_{\tau \leq t} f(x_\tau)$. Respectively, the cumulative inference regret is $R_t(k) = \sum_{\tau \leq t} f(x^*) - \max_{x} \mu_{\tau-1}(x)$. 
The algorithm converges to the optimum using all three kernels. The meta-learned kernel, however, improves upon using $\kfull$ and results in a performance competitive to when $\tk$ is known by \textsc{GP-UCB}. This behavior empirically confirms \cref{cor:regret_bound}.
\paragraph{Consistency Experiment} From \cref{cor:regret_bound} we concluded that, as the size of the meta-data grows, the regret bound achieved via $\khat$ converges to the oracle bound, i.e., the bound satisfied by the learner when it has knowledge of the true kernel.
As \cref{fig:bo_consistency} shows, this consistency is also reflected in the empirical regret values.
As we increase $m$, the number of offline tasks given to the meta-learner, the cumulative inference regret at $T=100$ improves. It approaches the regret obtained by the oracle algorithm. The value of $\lambda$ does not affect this convergence, as long as it satisfies \cref{thm:RKHS_recovery}. Similar to \cref{fig:regret_bo}, this plot is generated for a fixed random $\tk$, averaged over $50$ random instances of the meta-learning and BO problem. 

\begin{figure}[t]
    \centering
    \begin{subfigure}[b]{0.49\linewidth}
    \centering
    \includegraphics[width=\linewidth, trim=10 0 10 0, clip]{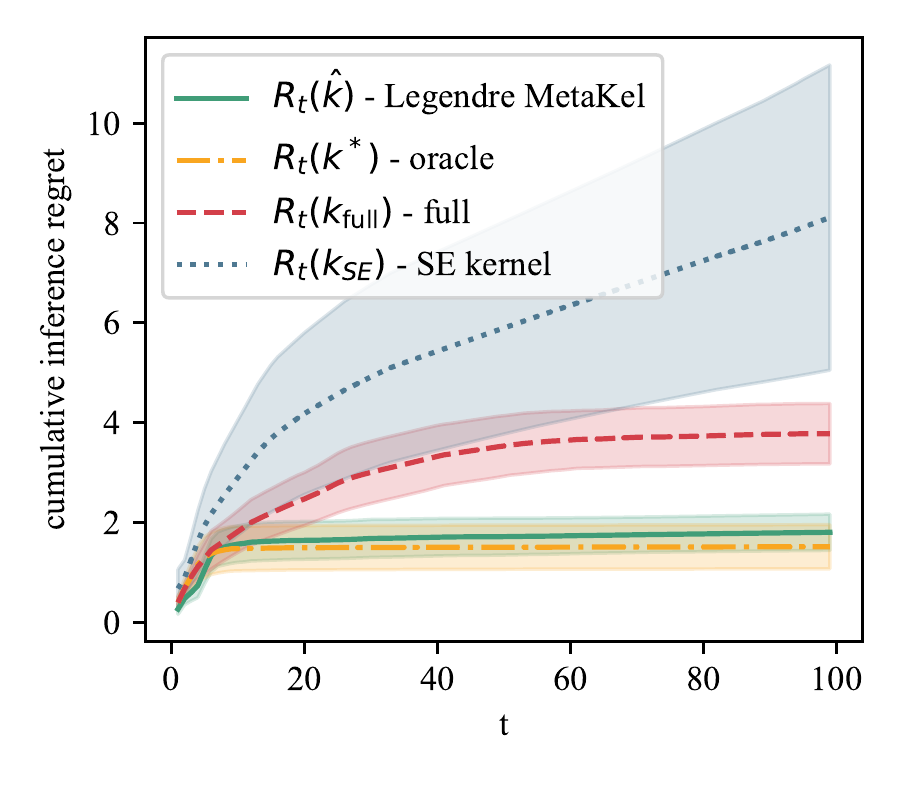}
    \vspaceexpfigure
    \end{subfigure}
    \hfill
    \begin{subfigure}[b]{0.49\linewidth}
    \includegraphics[width =\linewidth, trim=10 0 10 0, clip]{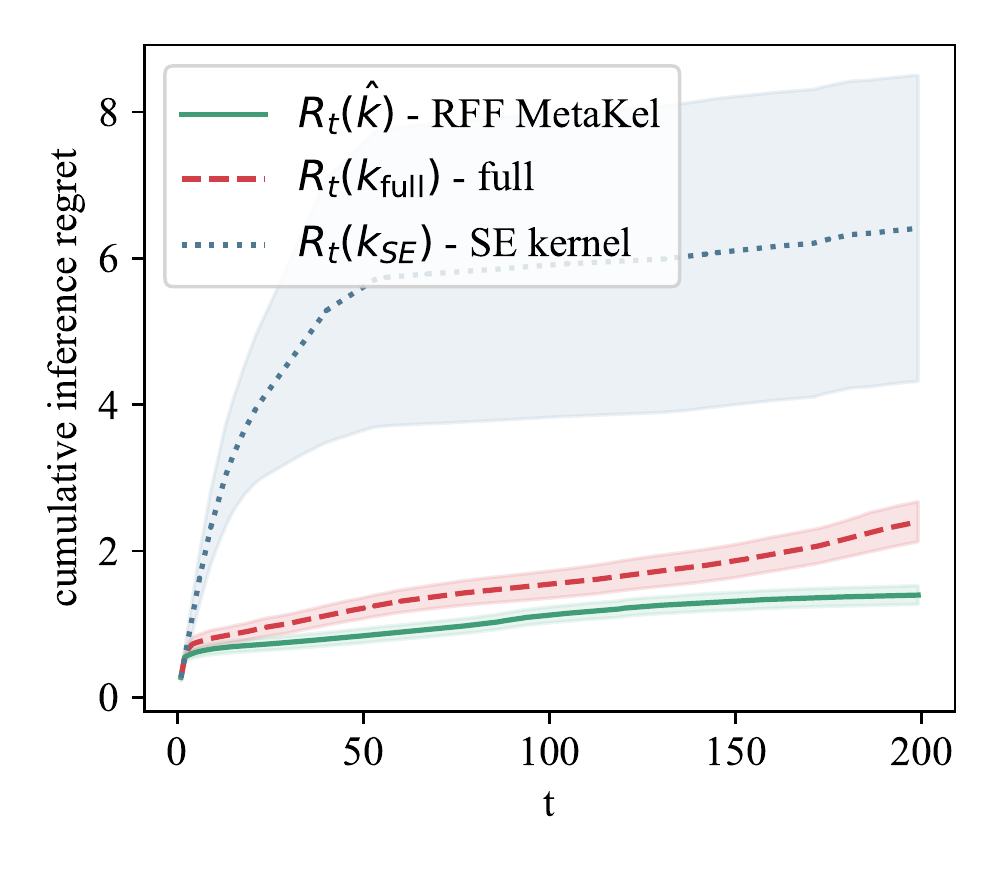}
    \vspaceexpfigure
    \end{subfigure}
    \vspace{-30pt}
    \caption{ \label{fig:regret_bo_2d}\looseness -1 Cumulative regret of GP-UCB. A synthetic 2D BO (left), hyper-parameters tuning of \textsc{GLMNET} (right).  \vspace{-12pt}}
\end{figure}
\vspaceparagraph
\paragraph{Regret Experiment for 2D domain} We repeat the regret experiment, with synthetic data over the $2$-dimensional domain $\gX = [-1,1]^2$. 
For $\vx = (x_1, x_2) \in \gX$, we define the Legendre feature map as
$
\vphi(\vx) = \left( P_j(x_1)P_{p-j}(x_2)\right)_{0 \leq j \leq p}
$. This feature map is $(p+1)$-dimensional, and has terms of degree at most $p$.
We use the polynomial $P_j(x_1)P_{p-j}(x_2)$ as the feature $\vphi_j(\vx)$ and create a synthetic dataset in a fashion identical to the $1$-dimensional case, again setting $p=20$ and $s=5$. \cref{fig:regret_bo_2d} shows the $2$-dimensional counterpart of \cref{fig:regret_bo}. 
Again, using $\khat$ results in competitive performance to the oracle algorithm with knowledge of $\tk$. Here, we also use \textsc{GP-UCB} together with the infinite-dimensional Squared Exponential (SE) kernel. The regret curve shows that we do not benefit from choosing a complex kernel for solving an inherently low dimensional problem.\looseness -1

\paragraph{Efficient Hyper-parameter Tuning with \mmkl} 
A common application of \textsc{GP-UCB} is in optimizing hyper-parameters of machine learning algorithms. In this setting, $\gX$ is the algorithm's hyper-parameter space, and $f$ represents the test performance of the algorithm. Evaluating each hyper-parameter configuration is costly, and thus the BO method has to be sample efficient.
We consider the \textsc{GLMNET} algorithm \citep{friedman2010regularization} and empirically demonstrate that by meta-learning the kernel, we gather knowledge from prior data, and in turn, tune the hyper-parameters of a new task more efficiently. Mainly, by running \textsc{GP-UCB} with $\khat$ we tend to find the optimal configuration of hyper-parameters for an unseen learning task faster, compared to using data-independent kernels (\cref{fig:regret_bo_2d}).
The OpenML platform \citep{Bischl2017OpenMLBS} enables access to data from hyper-parameter tuning of \textsc{GLMNET} on 38 different classification tasks. 
We split these datasets into a meta-dataset with $m=25$ and leave the rest as test tasks. 
For meta-learning the kernel, we use 500 Random Fourier Features \citep{rahimi2007random} defined on a $2$-dimensional domain, since the \textsc{GLMNET} algorithm has only two hyper-parameters.
The remaining details of our experiment setup is given in \cref{app:experiments}.
\cref{fig:regret_bo_2d} shows the performance of \textsc{GP-UCB} on the test task. Utilizing $\khat$ results in a sample-efficient \textsc{GP-UCB} that rapidly approaches an optimal choice of hyper-parameters. This is in contrast to running \textsc{GP-UCB} with the SE kernel, which takes about 50 iterations to find a good configuration. \looseness -1

\section{Conclusion}
We proposed \mmkl, a method for reliably learning kernel functions from offline meta-data. 
As a first, our approach yields provably valid meta-learned adaptive confidence sets, setting it apart from existing meta-learning approaches.
Importantly, we showed that our meta-learned kernel produces tight and consistent confidence bounds for the target function, provided that enough meta-data is available. As an example application, we showed that \textsc{GP-UCB} still yields sublinear regret when using the meta-learned kernel, performing competitively in theory and experiments with the oracle that has access to the unknown kernel.
We believe this result opens up avenues towards giving convergence guarantees for other sequential decision algorithms without oracle knowledge of the hypothesis space. 


\section*{Acknowledgements}
\looseness -1 We thank Andreea Musat and Victor Armegioiu for fruitful discussions and their contributions to an earlier research path of this project. In addition, we thank Scott Sussex and Anastasia Makarova for their feedback on the draft of this paper. We appreciate Felix Schur and Adrian M{\"u}ller's thorough feedback on the final manuscript. This research was supported by the European Research Council (ERC) under the European Union’s Horizon 2020 research and innovation program grant agreement no. 815943. Jonas Rothfuss was supported by an Apple Scholars in AI/ML fellowship.

\bibliographystyle{icml2022}
\bibliography{refs}

\begin{thebibliography}{57}
\providecommand{\natexlab}[1]{#1}
\providecommand{\url}[1]{\texttt{#1}}
\expandafter\ifx\csname urlstyle\endcsname\relax
  \providecommand{\doi}[1]{doi: #1}\else
  \providecommand{\doi}{doi: \begingroup \urlstyle{rm}\Url}\fi

\bibitem[Abbasi-Yadkori et~al.(2011)Abbasi-Yadkori, P\'{a}l, and
  Szepesv\'{a}ri]{abbasiyadkori2011}
Abbasi-Yadkori, Y., P\'{a}l, D., and Szepesv\'{a}ri, C.
\newblock Improved algorithms for linear stochastic bandits.
\newblock In \emph{Advances in Neural Information Processing Systems}, 2011.

\bibitem[Bach et~al.(2012)Bach, Jenatton, Mairal, Obozinski,
  et~al.]{bach2011optimization}
Bach, F., Jenatton, R., Mairal, J., Obozinski, G., et~al.
\newblock Optimization with sparsity-inducing penalties.
\newblock \emph{Foundations and Trends in Machine Learning}, 2012.

\bibitem[Bach(2008)]{bach2008consistency}
Bach, F.~R.
\newblock Consistency of the group lasso and multiple kernel learning.
\newblock \emph{Journal of Machine Learning Research}, 2008.

\bibitem[Bach et~al.(2004)Bach, Lanckriet, and Jordan]{bach2004multiple}
Bach, F.~R., Lanckriet, G.~R., and Jordan, M.~I.
\newblock Multiple kernel learning, conic duality, and the smo algorithm.
\newblock In \emph{Proceedings of the twenty-first international conference on
  Machine learning}, 2004.

\bibitem[Bastani \& Bayati(2020)Bastani and Bayati]{bastani2020online}
Bastani, H. and Bayati, M.
\newblock Online decision making with high-dimensional covariates.
\newblock \emph{Operations Research}, 2020.

\bibitem[Basu et~al.(2021)Basu, Kveton, Zaheer, and Szepesv{\'a}ri]{basu2021no}
Basu, S., Kveton, B., Zaheer, M., and Szepesv{\'a}ri, C.
\newblock No regrets for learning the prior in bandits.
\newblock \emph{Advances in Neural Information Processing Systems}, 34, 2021.

\bibitem[Berkenkamp et~al.(2017)Berkenkamp, Turchetta, Schoellig, and
  Krause]{berkenkamp2017safe}
Berkenkamp, F., Turchetta, M., Schoellig, A., and Krause, A.
\newblock Safe model-based reinforcement learning with stability guarantees.
\newblock \emph{Advances in neural information processing systems}, 2017.

\bibitem[Berkenkamp et~al.(2019)Berkenkamp, Schoellig, and
  Krause]{berkenkamp2019no}
Berkenkamp, F., Schoellig, A.~P., and Krause, A.
\newblock No-regret bayesian optimization with unknown hyperparameters.
\newblock \emph{Journal of Machine Learning Research}, 2019.

\bibitem[Bickel et~al.(2009)Bickel, Ritov, and
  Tsybakov]{bickel2009simultaneous}
Bickel, P.~J., Ritov, Y., and Tsybakov, A.~B.
\newblock Simultaneous analysis of lasso and dantzig selector.
\newblock \emph{The Annals of statistics}, 2009.

\bibitem[Bischl et~al.(2017)Bischl, Casalicchio, Feurer, Hutter, Lang,
  Mantovani, Rijn, and Vanschoren]{Bischl2017OpenMLBS}
Bischl, B., Casalicchio, G., Feurer, M., Hutter, F., Lang, M., Mantovani, R.,
  Rijn, J.~N., and Vanschoren, J.
\newblock Openml benchmarking suites and the openml100.
\newblock \emph{arXiv preprint arXiv:1708.03731}, 2017.

\bibitem[Bogunovic \& Krause(2021)Bogunovic and
  Krause]{bogunovic2021misspecified}
Bogunovic, I. and Krause, A.
\newblock Misspecified {G}aussian process bandit optimization.
\newblock In \emph{Conference on Neural Information Processing Systems
  (NeurIPS)}, 2021.

\bibitem[Boutilier et~al.(2020)Boutilier, Hsu, Kveton, Mladenov, Szepesvari,
  and Zaheer]{boutilier2020differentiable}
Boutilier, C., Hsu, C.-w., Kveton, B., Mladenov, M., Szepesvari, C., and
  Zaheer, M.
\newblock Differentiable meta-learning of bandit policies.
\newblock In \emph{Advances in Neural Information Processing Systems}, 2020.

\bibitem[Boyd et~al.(2004)Boyd, Boyd, and Vandenberghe]{boyd2004convex}
Boyd, S., Boyd, S.~P., and Vandenberghe, L.
\newblock \emph{Convex optimization}.
\newblock Cambridge University Press, 2004.

\bibitem[B{\"u}hlmann \& Van De~Geer(2011)B{\"u}hlmann and Van
  De~Geer]{buhlmann2011statistics}
B{\"u}hlmann, P. and Van De~Geer, S.
\newblock \emph{Statistics for high-dimensional data: methods, theory and
  applications}.
\newblock Springer Science \& Business Media, 2011.

\bibitem[Bunea et~al.(2013)Bunea, Lederer, and She]{bunea2013group}
Bunea, F., Lederer, J., and She, Y.
\newblock The group square-root lasso: Theoretical properties and fast
  algorithms.
\newblock \emph{IEEE Transactions on Information Theory}, 2013.

\bibitem[Cavalier et~al.(2002)Cavalier, Golubev, Picard, and
  Tsybakov]{cavalier2002oracle}
Cavalier, L., Golubev, G., Picard, D., and Tsybakov, A.
\newblock Oracle inequalities for inverse problems.
\newblock \emph{The Annals of Statistics}, 2002.

\bibitem[Cella \& Pontil(2021)Cella and Pontil]{cella2021multi}
Cella, L. and Pontil, M.
\newblock Multi-task and meta-learning with sparse linear bandits.
\newblock In \emph{Uncertainty in Artificial Intelligence}. PMLR, 2021.

\bibitem[Cella et~al.(2020)Cella, Lazaric, and Pontil]{cella2020meta}
Cella, L., Lazaric, A., and Pontil, M.
\newblock Meta-learning with stochastic linear bandits.
\newblock In \emph{International Conference on Machine Learning}. PMLR, 2020.

\bibitem[Chowdhury \& Gopalan(2017)Chowdhury and
  Gopalan]{chowdhury2017kernelized}
Chowdhury, S.~R. and Gopalan, A.
\newblock On kernelized multi-armed bandits.
\newblock In \emph{International Conference on Machine Learning}. PMLR, 2017.

\bibitem[Cristianini et~al.(2006)Cristianini, Kandola, Elisseeff, and
  Shawe-Taylor]{cristianini2006kernel}
Cristianini, N., Kandola, J., Elisseeff, A., and Shawe-Taylor, J.
\newblock On kernel target alignment.
\newblock In \emph{Innovations in machine learning}. Springer, 2006.

\bibitem[Curi et~al.(2020)Curi, Berkenkamp, and Krause]{curi2020efficient}
Curi, S., Berkenkamp, F., and Krause, A.
\newblock Efficient model-based reinforcement learning through optimistic
  policy search and planning.
\newblock \emph{Advances in Neural Information Processing Systems}, 2020.

\bibitem[Evgeniou \& Pontil(2004)Evgeniou and Pontil]{evgeniou2004regularized}
Evgeniou, T. and Pontil, M.
\newblock Regularized multi--task learning.
\newblock In \emph{Proceedings of the tenth ACM SIGKDD international conference
  on Knowledge discovery and data mining}, 2004.

\bibitem[Friedman et~al.(2010)Friedman, Hastie, and
  Tibshirani]{friedman2010regularization}
Friedman, J., Hastie, T., and Tibshirani, R.
\newblock Regularization paths for generalized linear models via coordinate
  descent.
\newblock \emph{Journal of statistical software}, 2010.

\bibitem[Gneiting et~al.(2007)Gneiting, Balabdaoui, and
  Raftery]{gneiting2007probabilistic}
Gneiting, T., Balabdaoui, F., and Raftery, A.~E.
\newblock Probabilistic forecasts, calibration and sharpness.
\newblock \emph{Journal of the Royal Statistical Society: Series B (Statistical
  Methodology)}, 2007.

\bibitem[G{\"o}nen \& Alpayd{\i}n(2011)G{\"o}nen and
  Alpayd{\i}n]{gonen2011multiple}
G{\"o}nen, M. and Alpayd{\i}n, E.
\newblock Multiple kernel learning algorithms.
\newblock \emph{The Journal of Machine Learning Research}, 2011.

\bibitem[Hao et~al.(2020)Hao, Lattimore, and Wang]{hao2020high}
Hao, B., Lattimore, T., and Wang, M.
\newblock High-dimensional sparse linear bandits.
\newblock \emph{Advances in Neural Information Processing Systems}, 2020.

\bibitem[Jacot et~al.(2018)Jacot, Gabriel, and Hongler]{jacot2018neural}
Jacot, A., Gabriel, F., and Hongler, C.
\newblock Neural tangent kernel: Convergence and generalization in neural
  networks.
\newblock \emph{Advances in neural information processing systems}, 2018.

\bibitem[Javanmard \& Montanari(2014)Javanmard and
  Montanari]{javanmard2014confidence}
Javanmard, A. and Montanari, A.
\newblock Confidence intervals and hypothesis testing for high-dimensional
  regression.
\newblock \emph{The Journal of Machine Learning Research}, 2014.

\bibitem[Kakade et~al.(2020)Kakade, Krishnamurthy, Lowrey, Ohnishi, and
  Sun]{kakade2020information}
Kakade, S., Krishnamurthy, A., Lowrey, K., Ohnishi, M., and Sun, W.
\newblock Information theoretic regret bounds for online nonlinear control.
\newblock \emph{Advances in Neural Information Processing Systems}, 2020.

\bibitem[Kim \& Paik(2019)Kim and Paik]{kim2019doubly}
Kim, G.-S. and Paik, M.~C.
\newblock Doubly-robust lasso bandit.
\newblock \emph{Advances in Neural Information Processing Systems}, 2019.

\bibitem[Kloft et~al.(2011)Kloft, Brefeld, Sonnenburg, and Zien]{kloft2011lp}
Kloft, M., Brefeld, U., Sonnenburg, S., and Zien, A.
\newblock Lp-norm multiple kernel learning.
\newblock \emph{The Journal of Machine Learning Research}, 2011.

\bibitem[Koltchinskii \& Yuan(2008)Koltchinskii and
  Yuan]{koltchinskii2008sparse}
Koltchinskii, V. and Yuan, M.
\newblock Sparse recovery in large ensembles of kernel machines.
\newblock In \emph{Proceedings of Conference on Learning Theory}, 2008.

\bibitem[K{\"u}hn et~al.(2018)K{\"u}hn, Probst, Thomas, and
  Bischl]{kuhn2018automatic}
K{\"u}hn, D., Probst, P., Thomas, J., and Bischl, B.
\newblock Automatic exploration of machine learning experiments on openml.
\newblock \emph{arXiv preprint arXiv:1806.10961}, 2018.

\bibitem[Kveton et~al.(2020)Kveton, Mladenov, Hsu, Zaheer, Szepesvari, and
  Boutilier]{kveton2020meta}
Kveton, B., Mladenov, M., Hsu, C.-W., Zaheer, M., Szepesvari, C., and
  Boutilier, C.
\newblock Meta-learning bandit policies by gradient ascent.
\newblock \emph{arXiv preprint arXiv:2006.05094}, 2020.

\bibitem[Liu \& Zhang(2009)Liu and Zhang]{liu2009estimation}
Liu, H. and Zhang, J.
\newblock Estimation consistency of the group lasso and its applications.
\newblock In \emph{Artificial Intelligence and Statistics}. PMLR, 2009.

\bibitem[Lounici et~al.(2011)Lounici, Pontil, Van De~Geer, and
  Tsybakov]{lounici2011oracle}
Lounici, K., Pontil, M., Van De~Geer, S., and Tsybakov, A.~B.
\newblock Oracle inequalities and optimal inference under group sparsity.
\newblock \emph{The annals of statistics}, 2011.

\bibitem[Massias et~al.(2018)Massias, Gramfort, and Salmon]{celer2018}
Massias, M., Gramfort, A., and Salmon, J.
\newblock Celer: a fast solver for the lasso with dual extrapolation.
\newblock In \emph{Proceedings of the 35th International Conference on Machine
  Learning}, 2018.

\bibitem[Ong et~al.(2005)Ong, Smola, and Williamson]{ong2005learning}
Ong, C.~S., Smola, A.~J., and Williamson, R.~C.
\newblock Learning the kernel with hyperkernels.
\newblock \emph{Journal of Machine Learning Research}, 2005.

\bibitem[Perrone et~al.(2008)Perrone, Jenatton, Seeger, and
  Archambeau]{perrone2018}
Perrone, V., Jenatton, R., Seeger, M.~W., and Archambeau, C.
\newblock {Scalable Hyperparameter Transfer Learning}.
\newblock In \emph{Advances in Neural Information Processing Systems}, 2008.

\bibitem[Rahimi et~al.(2007)Rahimi, Recht, et~al.]{rahimi2007random}
Rahimi, A., Recht, B., et~al.
\newblock Random features for large-scale kernel machines.
\newblock In \emph{NIPS}, 2007.

\bibitem[Rothfuss et~al.(2021{\natexlab{a}})Rothfuss, Fortuin, Josifoski, and
  Krause]{rothfuss2021pacoh}
Rothfuss, J., Fortuin, V., Josifoski, M., and Krause, A.
\newblock Pacoh: Bayes-optimal meta-learning with pac-guarantees.
\newblock In \emph{International Conference on Machine Learning}. PMLR,
  2021{\natexlab{a}}.

\bibitem[Rothfuss et~al.(2021{\natexlab{b}})Rothfuss, Heyn, Chen, and
  Krause]{rothfuss2021meta}
Rothfuss, J., Heyn, D., Chen, J., and Krause, A.
\newblock Meta-learning reliable priors in the function space.
\newblock In \emph{Advances in Neural Information Processing Systems},
  2021{\natexlab{b}}.

\bibitem[Russo \& Van~Roy(2014)Russo and Van~Roy]{russo2014learning}
Russo, D. and Van~Roy, B.
\newblock Learning to optimize via posterior sampling.
\newblock \emph{Mathematics of Operations Research}, 2014.

\bibitem[Sessa et~al.(2020)Sessa, Bogunovic, Kamgarpour, and
  Krause]{sessa2020learning}
Sessa, P.~G., Bogunovic, I., Kamgarpour, M., and Krause, A.
\newblock Learning to play sequential games versus unknown opponents.
\newblock \emph{Advances in Neural Information Processing Systems}, 2020.

\bibitem[Simchowitz et~al.(2021)Simchowitz, Tosh, Krishnamurthy, Hsu, Lykouris,
  Dudik, and Schapire]{simchowitz2021bayesian}
Simchowitz, M., Tosh, C., Krishnamurthy, A., Hsu, D.~J., Lykouris, T., Dudik,
  M., and Schapire, R.~E.
\newblock Bayesian decision-making under misspecified priors with applications
  to meta-learning.
\newblock \emph{Advances in Neural Information Processing Systems}, 2021.

\bibitem[Srinivas et~al.(2010)Srinivas, Krause, Kakade, and
  Seeger]{srinivas2009gaussian}
Srinivas, N., Krause, A., Kakade, S., and Seeger, M.
\newblock Gaussian process optimization in the bandit setting: No regret and
  experimental design.
\newblock In \emph{Proceedings of the 27th International Conference on
  International Conference on Machine Learning}, 2010.

\bibitem[Vakili et~al.(2021)Vakili, Khezeli, and
  Picheny]{vakili2021information}
Vakili, S., Khezeli, K., and Picheny, V.
\newblock On information gain and regret bounds in gaussian process bandits.
\newblock In \emph{International Conference on Artificial Intelligence and
  Statistics}. PMLR, 2021.

\bibitem[Van~de Geer et~al.(2011)Van~de Geer, B{\"u}hlmann, and
  Zhou]{van2011adaptive}
Van~de Geer, S., B{\"u}hlmann, P., and Zhou, S.
\newblock The adaptive and the thresholded lasso for potentially misspecified
  models (and a lower bound for the lasso).
\newblock \emph{Electronic Journal of Statistics}, 2011.

\bibitem[Vershynin(2018)]{vershynin2018high}
Vershynin, R.
\newblock \emph{High-dimensional probability: An introduction with applications
  in data science}, volume~47.
\newblock Cambridge University Press, 2018.

\bibitem[Wainwright(2019)]{wainwright2019high}
Wainwright, M.~J.
\newblock \emph{High-dimensional statistics: A non-asymptotic viewpoint}.
\newblock Cambridge University Press, 2019.

\bibitem[Wang et~al.(2018{\natexlab{a}})Wang, Wei, and Yao]{wang2018minimax}
Wang, X., Wei, M., and Yao, T.
\newblock Minimax concave penalized multi-armed bandit model with
  high-dimensional covariates.
\newblock In \emph{International Conference on Machine Learning}. PMLR,
  2018{\natexlab{a}}.

\bibitem[Wang \& de~Freitas(2014)Wang and de~Freitas]{wang2014theoretical}
Wang, Z. and de~Freitas, N.
\newblock Theoretical analysis of bayesian optimisation with unknown gaussian
  process hyper-parameters.
\newblock \emph{arXiv preprint arXiv:1406.7758}, 2014.

\bibitem[Wang et~al.(2017)Wang, Li, Jegelka, and Kohli]{wang2017batched}
Wang, Z., Li, C., Jegelka, S., and Kohli, P.
\newblock Batched high-dimensional bayesian optimization via structural kernel
  learning.
\newblock In \emph{International Conference on Machine Learning}. PMLR, 2017.

\bibitem[Wang et~al.(2018{\natexlab{b}})Wang, Kim, and
  Kaelbling]{wang2018regret}
Wang, Z., Kim, B., and Kaelbling, L.~P.
\newblock Regret bounds for meta bayesian optimization with an unknown gaussian
  process prior.
\newblock \emph{arXiv preprint arXiv:1811.09558}, 2018{\natexlab{b}}.

\bibitem[Wynne et~al.(2021)Wynne, Briol, and Girolami]{wynne2021convergence}
Wynne, G., Briol, F.-X., and Girolami, M.
\newblock Convergence guarantees for gaussian process means with misspecified
  likelihoods and smoothness.
\newblock \emph{Journal of Machine Learning Research}, 2021.

\bibitem[Zhao \& Yu(2006)Zhao and Yu]{zhao2006model}
Zhao, P. and Yu, B.
\newblock On model selection consistency of lasso.
\newblock \emph{The Journal of Machine Learning Research}, 2006.

\bibitem[Zhou et~al.(2020)Zhou, Li, and Gu]{zhou2020neural}
Zhou, D., Li, L., and Gu, Q.
\newblock Neural contextual bandits with ucb-based exploration.
\newblock In \emph{International Conference on Machine Learning}. PMLR, 2020.

\end{thebibliography}

\newpage
\appendix
\onecolumn
\numberwithin{equation}{section}
\icmltitle{Meta-Learning Hypothesis Spaces for Sequential Decision-making:\\ Supplementary Material}
\section{Details of the main Result}
\begin{table}[ht]
    \centering
    \begin{tabular}{l|l}
    symbol & description\\
    \hline
    $\tk$ &  true (unknown) kernel function \\
    $\khat$ &  meta-learned kernel \\
    \hline 
    $p$ & total number of candidate base kernels \\
    $k_j$ & base kernels that construct $\tk$ and $\khat$, $1 \leq j \leq p$\\
    $\teta_j$ & coefficient of $k_j$ in construction of $k^*$, i.e., $\tk(\cdot, \cdot) = \sum_{j=1}^p \teta_j k_j(\cdot, \cdot)$\\
    $\etahat_j$& coefficient of $k_j$ in construction of $\khat$, i.e., $\khat(\cdot, \cdot) = \sum_{j=1}^p \etahat_j k_j(\cdot, \cdot)$\\
    $\tveta$ & the vector $\left( \teta_1, \cdots, \teta_j, \cdots, \teta_p\right) \in \sR^p$\\
    $\vetahat$ & the vector $\left( \etahat_1, \cdots, \etahat_j, \cdots, \etahat_p\right) \in \sR^p$\\
    $J_\tk$ & the set $\{ 1 \leq j \leq p,\, \text{s.t.}\, \teta_j \neq 0\}$\\
    $J_\khat$ & the set $\{ 1 \leq j \leq p,\, \text{s.t.}\, \etahat_j \neq 0\}$ \\
    \hline    
    $\vphi_j(\cdot)$ & feature map for $k_j$, i.e., $k_j(\vx, \vx') = \vphi^T_j(\vx) \vphi_j(\vx')$ \\
    $\vphi(\cdot)$ & $\left(\sqrt{\teta_1}\vphi_1^T(\vx),\cdots, \sqrt{\teta_p}\vphi_p^T(\vx)\right)^T$, i.e., feature map for $\tk$\\
    \hline
    $d_0$ & dimension of the input domain, $\gX \subset \sR^{d_0}$\\
    $d_j$ & dimension of a feature map $\vphi_j(\vx) \in \sR^{d_j}$\\
    $\dmax$ &  $\max_{1\leq j\leq p} d_j$\\
    $d$ & $\sum_{j=1}^p d_j$ \\
    $d^*$ &$\sum_{j\in J_\tk} d_j$ \\
    $\hat d$ & $\sum_{j\in J_\khat} d_j$ \\
    \hline
    $n$ &  number of samples in each task $s$ in the meta-data\\
    $m$ &  number of tasks in the meta-data\\
    $f_s$ & target function from task $s$ \\
    $\tvbeta_s$ & true coefficients vector for task $s$: $f_s(\vx) =  \vphi^T(\vx) \tvbeta_s $, in $\sR^d$  \\
    $\vbetahat_s$ & estimate of $\tvbeta_s$ \\
    $\tvbeta_s\gj$ & sub-vector of $\tvbeta_s$ corresponding to kernel $k_j$, in $\sR^{d_j}$ \\
    $\vbetahat_s\gj$ & estimate of $\tvbeta_s\gj$\\
    $\tvbeta$ &  $(\tvbeta_1{^T},\cdots,\tvbeta_m{^T})^T$, in $\sR^{md}$\\
    $\vbetahat$  & estimate of $\tvbeta$\\
    $\tvbeta\gj$ & sub-vector of $\tvbeta$ corresponding to kernel $k_j$, in $\sR^{md_j}$ \\
    $\vbetahat\gj$ & estimate of $\tvbeta\gj$ \\
    \end{tabular}
    \caption{Notation Guide}
    \label{tab:notations}
\end{table}

\begin{figure}[ht]
\centering
\begin{minipage}{0.6\textwidth}
\centering
       \begin{minipage}{\textwidth}
    \centering
\[
\left(\begin{array}{c}
        \\
      \\
      \vy_s\\
      \\
      \\ 
\end{array}\right)
= \sum_{j=1}^p\left( \begin{array}{*5{c}}
     \tikzmark{left}{}&  &\vphi^T_j(\vx_{s,1})& &\\
     &&&&\\
     &&\vdots&&\\
     &&&&\\
     & & \vphi^T_j(\vx_{s,n})& & \tikzmark{right}{}\\
\end{array}\right)\Highlight \left(\begin{array}{c}
       \\
      \\
      \vbeta^{(j)}_s\\
      \\
      \\ 
\end{array}\right)+ \left(\begin{array}{c}
       \\
      \\
      \bm{\epsilon}_s\\
      \\
      \\ 
\end{array}\right)
\]
\end{minipage}
\begin{minipage}{\textwidth}
    \centering
 \begin{tikzpicture}[
    pmat/.style={
       matrix of math nodes,
       nodes={font=\footnotesize\strut, inner sep=1.5pt},
       left delimiter=(,
       right delimiter=),
       }
       ]
\matrix (y) [pmat]{
\vy_1\\
\vdots\\
\vy_s\\
\vdots\\
\vy_m\\
}; 
\matrix (phi)[pmat, right =40pt of y] {
        \mPhi_1 &   &         &   &        \\
                &   &         & \quad 0 &        \\
                &   & \mPhi_s &   &        \\
                & 0\quad &         &   &        \\
                &   &         &   & \mPhi_m\\
        } ;
        
\matrix (beta) [pmat, right =25pt of phi]{
\vbeta_1\\
\vdots\\
\vbeta_s\\
\vdots\\
\vbeta_m\\
}; 
\matrix (epsilon) [pmat, right =of beta]{
\bm\epsilon_1\\
\vdots\\
\bm\epsilon_s\\
\vdots\\
\bm\epsilon_m\\
}; 
        \draw [DG] (phi-1-1.center) to (phi-5-5.center);
        \path (y) -- node {$=$} (phi);
        \path (beta) -- node {$+$} (epsilon);
                
    \end{tikzpicture}
\end{minipage}
\end{minipage}
\begin{minipage}{0.2\textwidth}
    \begin{tikzpicture}[thick,scale=0.6, every node/.style={scale=0.95},
       pmat/.style={
       matrix of math nodes,
       nodes={font=\footnotesize\strut, inner sep=0.1pt},
       left delimiter=(,
       right delimiter=),
       }
       ]
\matrix (beta) [pmat]{
\vbeta^{(1)}_1 \\ \vdots \\ \vbeta^{(j)}_1 \\ \vdots \\ \vbeta_1^{(p)}\\ \vdots \\
\vbeta^{(1)}_s \\ \vdots \\ \vbeta^{(j)}_s \\ \vdots \\ \vbeta_s^{(p)}\\ \vdots \\ 
\vbeta^{(1)}_m \\ \vdots \\ \vbeta^{(j)}_m \\ \vdots \\ \vbeta_m^{(p)} \\
}; 
\matrix (beta2) [pmat, right =50 pt of beta]{
\vbeta^{(j)}_1 \\ \vdots \\ \vbeta^{(j)}_s  \\ \vdots \\ \vbeta^{(j)}_m\\ 
};
\draw [DGG] (beta-3-1.north) to (beta-3-1.south);
\draw[color=comment,opacity=.4,line width=1.5,inner sep=20pt, rounded corners] (beta-1-1.north west) rectangle (beta-5-1.south east);
\draw [DGG] (beta-9-1.north) to (beta-9-1.south);
\draw[color=comment,opacity=.4,line width=1.5,inner sep=20pt, rounded corners] (beta-7-1.north west) rectangle (beta-11-1.south east);
\draw [DGG] (beta-15-1.north) to (beta-15-1.south);
\draw [DGG] (beta2-1-1.north) to (beta2-5-1.south);
\draw[color=comment,opacity=.4,line width=1.5,inner sep=20pt,rounded corners] (beta-13-1.north west) rectangle (beta-17-1.south east);

 \node [left =5pt of beta] {$\vbeta = $};
\node [left = 5pt of beta2] {$\vbeta^{(j)} = $};
\end{tikzpicture}
\end{minipage}
\caption{\label{fig:glasso_formulation}Visual guide for the Group Lasso formulation. The red shade shows the group of coefficients which correspond to the effect of kernel $k_j$. The green shade demonstrates how features from each task come together on the diagonal of the multi-task feature matrix. The coefficients that belong to one task are group together with a green rectangle.}
\end{figure}

\subsection{RKHS Refresher} \label{app:rkhs}
Here we present a compact reminder of RKHS basics for the sake of completeness and clarifying our notation. We work in a finite-dimensional regime which can also be described by a euclidean vector space. Nevertheless, we use the RKHS notation as it gives a powerful framework and hides away the vector algebra.
This section is mainly based on \citet{wainwright2019high}. For a positive semi-definite kernel function $k$ over some set $\mathcal{X}\times \mathcal{X}$, the corresponding unique Reproducing Kernel Hilbert Space can be constructed as,
\[
\mathcal{H}_k = \Big\{f: \mathcal{X}\rightarrow \mathbb{R}\,\vert \, f(\cdot) = \sum_{i=1}^n \alpha_i k(\vx_i,\cdot), \, n \in \mathbb{N}, \, (\vx_i)_{i=1}^n\in \mathcal{X}, \bm{\alpha}\in \mathbb{R}^n  \Big\},
\]
equipped with the dot product, $\langle f, \bar{f}\rangle_k = \sum_{i,j} \alpha_i \bar{\alpha}_j k(x_i, \bar{x}_j)$. We limit $\mathcal{X}$ to compact sets, and only consider Mercer kernels, i.e., continuous kernel functions that satisfy the Hilbert-Schmidt condition, 
\[
\int_{\mathcal{X}\times\mathcal{X}} k^2(\vx,\vx')d\mu(\vx)d\mu(\vx') < \infty
\]
where $\mu$ is a non-negative measure over $\mathcal{X}$. Mercer's theorem states that under these assumptions, the kernel operator has a sequence of orthonormal eigenfunctions $(\phi_r)_{r\geq 1}$and non-negative eigenvalues $(\eta_r)_{r\geq 1}$, defined as follows
\begin{equation} \label{eq:eigenvalues}
\int_{\mathcal{X}} k(\vx,\vx')\phi_r(\vx') d\mu(\vx') = \eta_r \phi_r(\vx).
\end{equation}
Moreover, $k$ can be written as their linear combination,
\[
k(\vx,\vx') = \sum_r \eta_r \phi_r(\vx) \phi_r(\vx').
\]
Or as a non-negative combination of base kernels, $k_r(\vx,\vx') = \phi_r(\vx) \phi_r(\vx')$
\[
k(\vx,\vx') = \sum_r \eta_r k_r(\vx,\vx').
\]
It immediately follows that the unique RKHS corresponding to $k$ takes the form,
\[
\mathcal{H}_k = \left\{f:\gX\rightarrow\sR\,\vert \, f(\cdot) = \sum_{r\geq 1} \beta_r \phi_r(\cdot), \, \sum_{r: \eta_r \neq 0}\frac{\beta_r^2}{\eta_r} < \infty  \right\}
\]
and the inner product the following form,
\[
\langle f, g \rangle_k = \sum_{r:\eta_r\neq 0} \frac{\langle f, \phi_r\rangle_2 \langle g, \phi_r\rangle_2}{\eta_r},
\]
where $\langle \cdot, \cdot \rangle_2$ denotes the inner product in the $L^2(\mathcal{X})$. It is then implied that $\norm{f}_k = \sum_{r: \eta_r \neq 0}\beta_r^2/\eta_r$. 
Lastly, we define $\bm{\phi}(\vx)=\left(\sqrt{\eta_r}\phi_r(\vx)\right)_{r\geq 1}$ to be the feature map corresponding to $k$. In this paper we refer to the number of non-zero eigen-values as the dimension of the kernel. Note that under this convention, most kernel functions used in practice, e.g. RBF kernel or the Mat{\'e}rn family, are infinite-dimensional. 

\subsection{Proof of Proposition \ref{prop:eta_hat}}\label{app:eta_trick_proof}
By \cref{eq:f_decompose} we may write a parametric equivalent of Problem (\ref{eq:nonparam_opt}) in terms of the feature maps $\vphi_j$,
\begin{equation}\label{eq:generic_opt}
\begin{split}
    \min_{\substack{0\leq \veta,\\ \forall s:\vbeta_s}}\, \frac{1}{m} \sum_{s=1}^m &\left[ \frac{1}{n}\sum_{i=1}^n \left( y_{s,i} - \sum_{j=1}^p \sqrt{\eta_j}\vphi_j^T(\vx_{s,i})\vbeta_s^{(j)}\right)^2\right] \\
& + \frac{\lambda}{2} \sum_{s=1}^m\sum_{j=1}^p\norm{\vbeta_s^{(j)}}_2^2 + \frac{\lambda}{2}\norm{\veta}_1.
\end{split}
\end{equation}
This problem is jointly convex in $\veta$ and $(\vbeta_s)_{s\leq m}$, and has an optimal solution \citep{kloft2011lp}. 
Renaming the variable $\vbeta_s^{(j)} \leftarrow \vbeta_s^{(j)}/\sqrt{\eta_j}$ gives the following equivalent problem,
\[
\min_{\substack{0\leq \veta,\\ \forall s:\vbeta_s}} \frac{1}{m} \sum_{s=1}^m \left[ \frac{1}{n}\sum_{i=1}^n \left( y_{s,i} - \sum_{j=1}^p \vphi_j^T(\vx_{s,i})\vbeta_s^{(j)}\right)^2\right] + \frac{\lambda}{2} \sum_{j=1}^p\frac{\sum_{s=1}^m\norm{\vbeta_s^{(j)}}_2^2}{\eta_j} + \frac{\lambda}{2}\norm{\veta}_1.
\]
There is no constraint connecting the two variables, and renaming $\vbeta_s^{(j)}$ does not effect the optimization problem with respect to $\eta$. Therefore, $\vetahat$ is also an optima for this problem. Let $(\vetahat, \vbetahat)$ denote the solution to the problem above. 
We show that $\etahat$ has a closed form expression in terms of $\vbetahat$. This observation allows us to reduce the joint optimization problem to an equivalent problem which is only over $(\vbeta_s)$. We use the \emph{$\eta$-trick} introduced in  \citet{bach2004multiple}. The authors observe that for any two scalar variables $w$ and $v$,
\[
\abs{w} = \min_{v\geq 0} \frac{w^2}{2v} + \frac{v}{2},
\]
and $\hat v = \abs{w}$. Applying this trick with $w = \norm{\vbeta^{(j)}}_2$ and $v = \eta_j$ for all $j \leq p$ gives
\[
 \min_{\veta\geq 0} \frac{\lambda}{2}\sum_{j=1}^p \frac{\norm{\vbeta^{(j)}}_2^2}{\eta_j} + \frac{\lambda}{2}\norm{\veta}_1 = \lambda \sum_{j=1}^p\norm{\vbeta^{(j)}}_2,
\]
which results the following equivalent problem
\begin{equation}\label{eq:multi_loss_raw}
\min_{\forall s:\vbeta_s} \frac{1}{m} \sum_{s=1}^m \left[ \frac{1}{n}\sum_{i=1}^n \left( y_{s,i} - \sum_{j=1}^p \vphi_j^T(\vx_{s,i})\vbeta_s^{(j)}\right)^2\right] + \lambda \sum_{j=1}^p\sqrt{\sum_{s=1}^m\norm{\vbeta_s^{(j)}}_2^2}.
\end{equation}
Note that By definition of $\vbeta^{(j)}$, the second term satisfies $ \sum_{j=1}^p\sqrt{\sum_{s=1}^m\norm{\vbeta_s^{(j)}}_2^2} = \norm{\vbeta^{(j)}}_2$. Finally, by simply using the vectorized notation we get \cref{eq:glasso_opt_multi}, concluding the proof. 

\section{Proof of Statements in Section \ref{sec:meta}}


For the first three lemmas in this section we follow the technique in \citet{lounici2011oracle} and occasionally use classical ideas established in \citet{buhlmann2011statistics}.

\paragraph{Notations and naming conventions} When $X$ is a matrix, $\norm{X}_2$ and $\norm{X}_{\mathrm{F}}$ denote its spectral and Frobenius norm, respectively. Consider the multi-task coefficients vector $\vbeta \in \sR^{md}$, and the sub-vector $\vbeta\gj \in \sR^{md_j}$ which denotes the coefficients corresponding to kernel $k_j$. Through out this proof, we use the convention ``group $j$" to refer to the set of indices of $\vbeta$ which indicate $\vbeta\gj$. Similarly, we let $\mPhi\gj$ denote the $mn\times md_j$ sub-matrix which only has the features coming from group $j$.  Lastly, let $\mPsi := \mPhi^T\mPhi/mn$, then $\mPsi\gj = (\mPhi\gj)^T\mPhi\gj/mn$ indicates the $md_j\times md_j$ submatrix that is caused by group $j$. 

\subsection{Proof of \cref{thm:RKHS_recovery}} \label{app:proof_rkhs}
Recall the vectorized formulation of the \mmkl loss,
\[
\gL(\vbeta) =  \frac{1}{mn} \norm{\vy-\mPhi\vbeta}_2^2 + \lambda \sum_{j=1}^p \norm{\vbeta\gj}_2.
\]
Let $\vvarepsilon_s = (\varepsilon_{s,i})_{i\leq n}$ denote error for task $s$ and $\vvarepsilon \in \sR^{mn}$ the stacked multi-task error vector. Using $\vy = \mPhi\vbeta + \vvarepsilon$, we may decompose the loss into two deterministic and random parts.
The term $2\vvarepsilon^T\mPhi(\vbetahat-\tvbeta)/mn$ is the random one and we will refer to as the empirical process. The first typical step in bounding the estimation error of Lasso estimators, is showing that the empirical process, which comes from the noise in observing values of $\vy$, does not play a drastic role.
More formally, let $A_j = \left\{ \norm{(\mPhi^T\vvarepsilon)\gj}_2/mn  \leq \lambda/4 \right\}$ denote that the event that the image of noise affecting the feature space of $\vphi_j$, is dominated by the regularization term of $\vbeta\gj$. In Lemma \ref{lem:A_happens} we show that $\cap_{j=1}^pA_j$ happens with high probability, if $\lambda$ is set properly. 
\begin{lemma}[Regularization term dominates the empirical process]\label{lem:A_happens}
Set $0<\delta<1$. Consider the random event $A = \cap_{j=1}^pA_j$. Then $A$ happens with probability greater than $1-\delta$, if
\[
\lambda \geq \frac{4\sigma}{\sqrt{mn}}\sqrt{1 + \frac{2}{m}\left( \log (2p/\delta) + \sqrt{m\dmax \log( 2p/\delta)}\right)}.
\]
where $\dmax = \max_{1\leq j\leq p} d_j$.
\end{lemma}
We now show that if the empirical process is controlled by regularization, i.e. if $\lambda$ is set to be large enough, then $\vbetahat = \min \gL(\vbeta)$ has favorable properties.
\begin{lemma}[Conditional properties of $\vbetahat$]\label{lem:prop_beta}
Assume that event $A$ happens. Then for any solution $\vbetahat$ of problem \ref{eq:glasso_opt_multi} and all $\vbeta \in \sR^{md}$, the following hold:
\begin{align}
    &\frac{1}{mn} \norm{\mPhi(\vbetahat - \tvbeta)}_2^2 + \frac{\lambda}{2} \sum_{j=1}^p \norm{\vbetahat\gj - \tvbeta\gj}_2 \leq \frac{1}{mn} \norm{\mPhi(\vbeta- \tvbeta)}_2^2 \notag\\
    & \quad\quad\quad+ 2\lambda \sum_{j:\vbeta\gj\neq 0} \min\left(\norm{\vbeta\gj}_2, \norm{\vbetahat\gj-\vbeta\gj}\right)\label{eq:prop1}\\
    & \left\vert \left\{j: \, \vbetahat\gj \neq 0 \right\}\right\vert \leq \frac{16}{(mn\lambda)^2}\norm{\mPhi(\vbetahat-\tvbeta)}_2^2\label{eq:prop3}
\end{align}
\end{lemma}
Note that by the meta data-generating model (Section \ref{sec:model}), $J_\tk =  \{j: \tvbeta\gj \neq 0\}$.
\begin{lemma}[Complete Variable Screening]\label{lem:var_screen} Assume $\vert J_\tk\vert \leq s$ and set $0 < \delta\leq 1$. Define
\[
\epsilon(n,m) = \frac{32\sigma s}{\kappa^2\sqrt{mn}}\sqrt{1 + \frac{2}{m}\left( \log (2p/\delta) + \sqrt{m\dmax \log( 2p/\delta)}\right)}  
\]
Under assumption \ref{cond:RE} with $\kappa=\kappa(s)$, if $\lambda$ is chosen according to lemma \ref{lem:A_happens}, then with probability greater than $1-\delta$
\begin{equation}\label{eq:scr_prop1}
\max_{j \in J_\tk} \norm{\vbetahat\gj-(\tvbeta)\gj}_2 \leq \epsilon(n,m)  
\end{equation}
and if in addition $\min_{j \in J_\tk} \norm{\tvbeta\gj}_2 \geq c_1$, then with the same probability for all $j \in J_\tk$
\begin{equation}\label{eq:scr_prop2}
\abs{\norm{\vbetahat\gj}_2 - c_1} \leq \epsilon(n,m).    
\end{equation}
\end{lemma}

We now turn to the first claim made in \cref{thm:RKHS_recovery}, and prove that $\gH_\tk \subseteq \Hhat$.
Since $\etahat_j \geq 0$ and $k_i$ are Mercer, then $\khat$ is also Mercer and corresponds to an RKHS which we have been referring to as $\Hhat$. 
Consider the RKHS $\tH$, since $\vert J_\tk\vert$ and each $d_j$s are finite,
\[
\tH = \left\{f: f(\cdot)= \sum_{j \in J_\tk} \sqrt{\teta_j}\vbeta_j^T\vphi_j(\cdot), \,\vbeta_j\in\sR^d,\,\norm{\vbeta_j} < \infty \right\}
\]
Therefore, $f \in \tH$ if and only if it is in the finite span of $\vphi$, defined as
\[
    \mathrm{FinSpan}\left(\{\vphi_j: j\in J_\tk\}\right) = \left\{f: f(\cdot)= \sum_{j \in J_\tk} \vbeta_j^T\vphi_j(\cdot), \,\vbeta_j \in\sR^d,\,\norm{\vbeta_j} < \infty \right\}
\]

Lemma \ref{lem:var_screen} states that  $\vbetahat\gj \geq c_1 - \epsilon(n,m)$ with probability greater than $1-\delta$ for $j \in J_\tk$. Therefore, for any $j \in J_\tk$, we get $\etahat_j \geq 1 - \epsilon(n,m)/c_1$, since we had set $\etahat_j = \norm{\vbetahat\gj}/c_1$. If $c_1 > \epsilon(n,m)$, then $\etahat_j >0$ and $j \in J_\khat$, implying $J_\tk \subset J_\khat$. 
Hence, under the assumptions of the theorem, with probability greater than $1-\delta$,
\[
 \tH = \mathrm{FinSpan}\left(\{\vphi_j: j\in J_\tk\}\right) \stackrel{\text{w.h.p.}}{\subset}  \mathrm{FinSpan}\left(\{\vphi_j: j\in J_\khat\}\right) = \Hhat.
\]
The next lemma bounds the $\khat$-norm of functions contained in $\tH$, concluding the proof for \cref{thm:RKHS_recovery}.

\begin{lemma}[Bounding the $\khat$-norm]\label{lem:norm_bound}
Set $0< \delta \leq 1$ and choose $\lambda$ according to Lemma \ref{lem:A_happens} and define $\epsilon(n,m)$ according to Lemma \ref{lem:var_screen}. If $\vert J_\tk \vert \leq s$ and Assumption \ref{cond:RE} holds with $\kappa = \kappa(s)$, then under Condition \ref{cond:betamin}, for all $f\in \tH$ with $\norm{f}_\tk \leq B$, 
\begin{equation}
    \norm{f}_\khat \leq \left(1+\frac{\epsilon(n,m)}{2c_1}+ o\left(\epsilon(n,m)\right)\right)
\end{equation}
\end{lemma}
\subsection{Proof of Proposition \ref{prop:sparsity_bound}}\label{proof:sparsity_bound}

Under assumptions of the proposition, event $A$ happens with probability greater than $1-\delta$.
From \cref{eq:prop3} in Lemma \ref{lem:prop_beta},
\[
\left\vert J_\khat \right\vert \leq \frac{16}{(mn\lambda)^2}\norm{\mPhi(\vbetahat-\tvbeta)}_2^2
\]
and by \cref{eq:aux3_beta_scr},
\[
    \frac{1}{\sqrt{mn}} \norm{\mPhi(\vbetahat-\tvbeta)}_2 \leq \frac{2\lambda\sqrt{s}}{\kappa}
\]
which gives
\[
\vert J_\khat \vert \leq \frac{64s}{mn\kappa^2(s)}
\]
Therefore, if $mn = \gO(s/p)$ then
$
\left\vert J_\khat \right\vert \leq p = \left\vert J_\kfull \right\vert
$ with probability greater than $1-\delta$.
and via similar argument as given in the proof of theorem \ref{thm:RKHS_recovery},
\[
 \Hhat = \mathrm{FinSpan}\left(\{\vphi_j: j\in J_\khat\}\right) \stackrel{\text{w.h.p.}}{\subset}  \mathrm{FinSpan}\left(\{\vphi_j: j\in J_\kfull\}\right) = \gH_\kfull.
\]

\subsection{Proof of Lemmas used in Section \ref{app:proof_rkhs}}
This section presents the proofs to the helper lemmas introduced before.
\begin{proof}[\textbf{Proof of Lemma \ref{lem:A_happens}}]\label{proof:A_happens}
This proof follows a similar treatment of the empirical process to Lemma 3.1 \citet{lounici2011oracle}.
Since $\varepsilon_{i,s}$ are i.i.d.~zero-mean sub-gaussian variables, we observe that
\[
\sP(A_j) = \sP\left( \left\{ \frac{1}{(mn)^2}\vvarepsilon^T\mPhi\gj(\mPhi\gj)^T\vvarepsilon \leq \frac{\lambda^2}{16}\right\}\right) = \sP\left(\left\{  
\frac{\sum_{i=1}^{mn}v_i(z_i^2-1)}{\sqrt{2}\norm{\vv}}\leq \alpha
\right\}\right)
\]
where $z_i$ are i.i.d. sub-gaussian variables with variance proxy $1$, $v_i$ denote the eigenvalues of $\mPhi\gj(\mPhi\gj)^T/mn$ and $\vv$ is the vector of these eigenvalues. Lastly,
\[
\alpha = \frac{mn\lambda^2/(16\sigma^2)-\tr(\mPsi\gj)}{\sqrt{2}\norm{\mPsi\gj}_{\mathrm{F}}}
\]
From Equation 27 \citet{cavalier2002oracle} yields the following inequality,
\[
\sP(A^c_j) = \sP\left(\left\vert \frac{\sum_{i=1}^{mn}(z_i^2-1)v_i}{\sqrt{2}\norm{\vv}_2}\right\vert > \alpha\right) \leq 2 \exp\left( -\frac{\alpha^2}{2(1+\sqrt{2}\alpha \norm{\vv}_\infty/\norm{\vv}_2)}\right)
\]
We choose $\lambda$ such that the right hand side is bounded by $\delta/p$.
By definition of $\vv$ we have $\norm{\vv}_\infty/\norm{\vv}_2 = \norm{\Psi\gj}_2/\norm{\Psi\gj}_{\mathrm{F}}$. Then, for $A^c_j$ to happen with probability smaller than $\delta/p$,
\[
\lambda \geq \frac{4\sigma}{\sqrt{mn}}\sqrt{\tr(\mPsi\gj) + 2\norm{\mPsi\gj}_2 \left ( 2\log (2p/\delta)+\sqrt{md_j\log(2p/\delta)}\right)}.
\]
Then by union bound, $A$ happens with probability greater than $1-\delta$ if
\[
\lambda \geq \max_j \frac{4\sigma}{\sqrt{mn}}\sqrt{\tr(\mPsi\gj) + 2\norm{\mPsi\gj}_2 \left ( 2\log (2p/\delta)+\sqrt{md_j\log(2p/\delta)}\right)}.
\]
Since the base kernels are normalized we may bound the norm and trace of $\mPsi\gj$.
\[
\tr(\mPsi\gj) = \frac{1}{mn}\sum_{s=1}^m \tr\left((\mPhi_s\gj)^T\mPhi_s\gj\right) = \frac{1}{mn}\sum_{s=1}^m\sum_{i=1}^n\vphi_j^T(\vx_{s,i})\vphi_j(\vx_{s,i}) = \frac{1}{mn}\sum_{s=1}^m\sum_{i=1}^n k_j(\vx_{s,i}, \vx_{s,i})\leq 1\]
Similarly, $\norm{\mPsi\gj}_2 \leq \frac{1}{mn}\max_s \sum_{i=1}^n k_j(\vx_{s,i},\vx_{s,i}) \leq 1/m$, and thereby concluding the proof. 
\end{proof}
\begin{proof}[\textbf{Proof of Lemma \ref{lem:prop_beta}}]
For proving this lemma we are essentially only using Cauchy-Schwarz and the Triangle inequality, together with the KKT optimality condition for $\gL$. For any $\vbeta$, since $\vbetahat$ is the minimizer of $\gL$ and due to the data generating model (Eq. \ref{eq:meta_data_model}), we have
\[
\frac{1}{mn} \norm{\mPhi(\vbetahat-\tvbeta)}_2^2 \leq  \frac{1}{mn} \norm{\mPhi(\vbeta-\tvbeta)}_2^2 + \frac{2}{mn}\vvarepsilon^T\mPhi(\vbetahat-\vbeta) + \lambda \sum_{j=1}^p \left(\norm{\vbeta\gj}_2 -\norm{\vbetahat\gj}_2\right).
\]
By Cauchy-Schwarz and the assumption that $A$ happens,
\begin{align*}
    \vvarepsilon^T\mPhi(\vbetahat-\vbeta) \leq \sum_{j=1}^p \norm{(\mPhi^T\vvarepsilon)\gj}_2\norm{\vbetahat\gj-\vbeta\gj}_2 \leq \frac{mn\lambda}{4} \sum_{j=1}^p \norm{\vbetahat\gj-\vbeta\gj}_2,
\end{align*}
and thereby,
\[
\frac{1}{mn} \norm{\mPhi(\vbetahat-\tvbeta)}_2^2 + \frac{\lambda}{2} \sum_{j=1}^p \norm{\vbetahat\gj-\vbeta\gj}_2 \leq  \frac{1}{mn} \norm{\mPhi(\vbeta-\tvbeta)}_2^2 + \lambda \sum_{j=1}^p \left(\norm{\vbetahat\gj-\vbeta\gj}_2 + \norm{\vbeta\gj}_2 -\norm{\vbetahat\gj}_2\right)
\]
which gives Inequality \ref{eq:prop1}.
By the KKT optimality conditions for convex losses \citep{boyd2004convex}, $\vbetahat$ is a minimizer of $\gL$, if and only if $0 \in \partial \gL(\vbetahat)$, where $\partial\gL(\vbetahat)$ denotes the sub-gradient of the loss evaluated at $\vbetahat$. Therefore $\vbetahat$ satisfies
\begin{align}
    \frac{2}{mn}\left( \mPhi^T(\vy-\mPhi\vbetahat) \right)\gj = \frac{\lambda\vbetahat\gj}{\norm{\vbetahat\gj}}, & \quad\text{if }\vbetahat\gj \neq 0\label{eq:kkt1}\\
    \frac{2}{mn}\norm{\left( \mPhi^T(\vy-\mPhi\vbetahat) \right)\gj}_2 \leq \lambda,  &\quad\text{if }\vbetahat\gj=0. \label{eq:kkt2}
\end{align}
As for Inequality \ref{eq:prop3}, conditioned on event $A$, by \cref{eq:meta_data_model} together with the KKT condition \ref{eq:kkt1}, we obtain that for all $1\leq j\leq p$ where $\vbetahat\gj\neq 0$,
\[
\frac{1}{mn}\norm{(\mPhi^T\mPhi(\vbetahat-\tvbeta))\gj}_2 \geq \frac{\lambda}{4}.
\]
Following the analysis of \citeauthor{lounici2011oracle} we conclude,
\begin{equation*}
\begin{split}
\left\vert \left\{j: \, \vbetahat\gj \neq 0 \right\}\right\vert & \leq
\frac{16}{(mn\lambda)^2}\sum_{j:\,\vbetahat\neq0} \norm{(\mPhi^T\mPhi(\vbetahat-\tvbeta))\gj}^2_2 \\
&\leq \frac{16}{(mn\lambda)^2}\sum_{j=1}^p \norm{(\mPhi^T\mPhi(\vbetahat-\tvbeta))\gj}^2_2 \\
& \leq \frac{16}{(mn\lambda)^2}\norm{(\mPhi^T\mPhi(\vbetahat-\tvbeta))}^2_2\\
& \leq \frac{16}{(mn\lambda)^2}\norm{(\mPhi(\vbetahat-\tvbeta))}^2_2.
\end{split}
\end{equation*}
Since the kernels $k_j$ are normalized by $1$ and $\norm{\mPhi}_2\leq \max_j k_j(\vx,\vx) \leq 1$.
\end{proof}
\begin{proof}[\textbf{Proof of Lemma \ref{lem:var_screen}}] Let $\vbeta\grJ= (\vbeta\gj)_{j\in J_\tk}$ denote the sub coefficient vector that corresponds to all the active groups. Due to \cref{eq:prop1} with $\vbeta = \tvbeta$,  conditioned on event $A$ we have
\begin{align}\label{eq:aux1_beta_scr}
    \frac{1}{mn} \norm{\mPhi(\vbetahat-\tvbeta)}_2^2\leq 2\lambda \sum_{j\in J_\tk} \norm{\vbetahat\gj-\tvbeta\gj} \leq 2\lambda \sqrt{\sum_{j\in J_\tk}s\norm{\vbetahat\gj-\tvbeta\gj}_2^2}=2\lambda \sqrt{s}\norm{\vbetahat\grJ-\tvbeta\grJ}_2
\end{align}
where the second inequality follows from Cauchy-Schwarz together with the Lemma's assumption $\vert J_\tk \vert \leq s$. By again using \cref{eq:prop1} we get $ \sum_{j=1}^p \norm{\vbetahat\gj-\tvbeta\gj}_2\leq 4 \sum_{j\in J_\tk} \norm{\vbetahat\gj-\tvbeta\gj}_2$, and thereby $ \sum_{j\notin J_\tk}^p \norm{\vbetahat\gj-\tvbeta\gj}_2\leq 3 \sum_{j\in J_\tk} \norm{\vbetahat\gj-\tvbeta\gj}_2$.
Using assumption \ref{cond:RE}, this inequality indicates that
\begin{equation}\label{eq:aux2_beta_scr}
   \norm{\vbetahat\grJ-\tvbeta\grJ}_2 \leq \frac{1}{\kappa \sqrt{mn}} \norm{\mPhi(\vbetahat-\tvbeta)}_2, 
\end{equation}
which together with \cref{eq:aux1_beta_scr} gives,
\begin{equation}\label{eq:aux3_beta_scr}
    \frac{1}{\sqrt{mn}} \norm{\mPhi(\vbetahat-\tvbeta)}_2 \leq \frac{2\lambda\sqrt{s}}{\kappa}
\end{equation}
The next chain of inequalities proves the first statement of the Lemma (\cref{eq:scr_prop1}). For all $j\in J_\tk$,
\begin{equation*}
    \begin{split}
        \norm{\vbetahat\gj-\tvbeta\gj}_2  \leq &\,\sum_{j=1}^p \norm{\vbetahat\gj-\tvbeta\gj}_2\\
         \stackrel{\text{\ref{eq:prop1}}}{\leq} & 4 \sum_{j\in J}  \norm{\vbetahat\gj-\tvbeta\gj}_2\\
         \stackrel{\text{\ref{eq:aux1_beta_scr}}}{\leq}&  4 \sqrt{s} \norm{\vbetahat\grJ - \tvbeta\grJ}_2\\
         \stackrel{\text{\ref{eq:aux2_beta_scr}}}{\leq} &  \frac{4\sqrt{s}}{\kappa \sqrt{mn}} \norm{\mPhi(\vbetahat-\tvbeta)}_2\\
         \stackrel{\text{\ref{eq:aux3_beta_scr}}}{\leq} & \,\frac{8\lambda s}{\kappa^2}
    \end{split}
\end{equation*}
Note that the analysis here is carried out conditional on event $A$. From Lemma \ref{lem:A_happens}, we have that $A$ happens with probability greater than $1-\delta$ if $\lambda$ is set according to the statement of the Lemma. Recall that from the Beta-min condition (Cond. \ref{cond:betamin}) we have $\norm{\tvbeta\gj}> c_1$, which together with \cref{eq:scr_prop1} concludes the proof.
\end{proof}
\begin{proof}[\textbf{Proof of Lemma \ref{lem:norm_bound}}]
Let $f \in \tH$, with $\norm{f}_k^2 \leq B^2$. Then by construction of $\tH$ (see Appendix \ref{app:rkhs})
\[f(\vx) = \sum_{j=1}^p \sqrt{\teta_j}\vphi_j^T(\vx)(\vbeta)\gj, \quad \norm{f}_k^2 = \sum_{j: \teta_j \neq 0}\norm{(\vbeta)\gj}_2^2 = \norm{\vbeta}_2^2.
\] 
We calculate the $\khat$ norm of functions that lie in $\tH$.  Let $I = \{1\leq j\leq p: \teta_j \neq 0, \etahat_j \neq 0\}$.
\begin{equation*}
    \norm{f}^2_{\khat} = \sum_{j: \etahat_j \neq 0}\sum_{r=1}^{d_j} \frac{\left( \langle f, \phi_{j,r} \rangle_{2} \right)^2}{\etahat_j} = \sum_{j: \etahat_j \neq 0}\sum_{r=1}^{d_j} \frac{\left( \sqrt{\teta_j}\vbeta_r\gj\right)^2}{\etahat_j} = \sum_{j \in I}\frac{\teta_j}{\etahat_j}\norm{\vbeta\gj}_2^2 = \sum_{j \in I}\frac{\teta_j}{\etahat_j}\norm{\vbeta\gj}_2^2
\end{equation*}
where $\phi_{j,r}$ denotes the $r$-th feature in the feature map $\vphi_j$, similarly $\vbeta_{r}\gj$ the $r$-th element in vector $\vbeta\gj$, and $\langle \cdot, \cdot \rangle_2$ is the inner product in the $L^2(\mathcal{X})$ space. By applying Cauchy-Schwarz we get
\[
    \norm{f}^2_{\khat} \leq \sqrt{\sum_{j \in I}(\frac{\teta_j}{\etahat_j})^2}\cdot\sqrt{\sum_{j \in I} \norm{\vbeta\gj}_2^4}.
 \]   
Consider the vector $\vv = (\norm{\vbeta\gj}_2^2)_{j\in I}$, we observe that $\norm{\vv}_2 = \sqrt{\sum_{j \in I} \norm{\vbeta\gj}_2^4}$ and $\norm{\vv}_1 \leq B^2$. Since $\norm{\cdot}_2 \leq \norm{\cdot}_1$ and due to the assumption $\norm{\tveta}_1\leq 1$, we obtain 
 \begin{equation}
    \norm{f}^2_\khat \leq B^2 \sum_{j \in I} \frac{\teta_j}{\etahat_j} \leq B^2 \norm{\tveta}_1 \max_{j \in I}\frac{1}{\etahat_j} \leq B^2 \max_{j \in I}\frac{1}{\etahat_j}.
\end{equation}
It remains to bound $\max_{j\in I}\etahat_j^{-1}$. We set $\vetahat_j = \norm{\vbetahat\gj}_2/(c1)$ and under conditions of the theorem, Lemma \ref{lem:var_screen} states that $\vetahat_j \geq 1-\epsilon(n,m)/c_1$, for all $j \in J_\tk$. Then for members of $I \subset J_\tk$,
\begin{equation*}
    \frac{1}{\etahat_j} \leq \frac{1}{1-\epsilon(n,m)/c_1} \leq \left( 1 + \frac{\epsilon(n,m) }{c_1}+ o\left(\epsilon(n,m)\right)\right),
\end{equation*}
which implies the following for the norm bound
\begin{equation}
    \norm{f}_\khat \leq B\left(1+\frac{\epsilon(n,m)}{2c_1}+ o\left(\epsilon(n,m)\right)\right)
\end{equation}
with probability greater than $1-\delta$.
\end{proof}
\section{Proof of Statements in Section \ref{sec:BO}}\label{app:BO}
The following lemma presents a confidence bound for when the learner has oracle knowledge of the true kernel. This lemma plays an integral role in for the proofs in this section.
\begin{lemma}[Theorem 2 \citet{chowdhury2017kernelized} for $\khat$]\label{lem:vanilla_CI}
Let $f \in \gH_\khat$ for some kernel $\hat k$ with a $\hat k$-norm bounded by $\hat B$.
Then with probability greater than $1-\bar\delta$, for all $\vx \in \gX$ and $t \geq 1$,
\begin{equation}
    \vert \hat \mu_{t-1}(\vx) - f(\vx)\vert \leq \hat \sigma_{t-1}(\vx)\left( \hat B + \sigma\sqrt{2(\hat \gamma_{t-1}+1+\log(1/\bar \delta))}\right)
\end{equation}
where $\hat \mu_{t-1}$ and $\hat \sigma_{t-1}$ are as defined in Equation \ref{eq:GPposteriors} with $\bar \sigma = 1 + 2/T$.
\end{lemma}
We skip the proof for this lemma as it is given in \citet{chowdhury2017kernelized}, with the same notation.
For the kernel $\hat k$ we define the \emph{maximum information gain} after $t-1$ observations as
\[\hat\gamma_{t-1} := \max_{[\vx_\tau]_{\tau \leq t}}\frac{1}{2}\log\det(\bm{I} + \bar\sigma^{-2}\hat\mK_{t-1})\]
This parameter quantifies the speed at which we learn about $f$, when using the kernel $\khat$. Note that $\gamma_{t-1}$ is independent of any specific realization of $H_{t-1}$. It only depends on the choice of kernel, the input domain, and the noise variance. The next lemma bounds this parameter.
\begin{lemma}[Information Gain Bound]\label{lem:infogain} The maximum information gain for $\khat$ after observing $t$ samples satisfies,
\[
\hat\gamma_{t}\leq \frac{\hat d}{2} \log\big(  1 +  \frac{\bar\sigma^{-2}t}{c_1} \big) = \gO(\hat d \log t/c_1)
\]
where $\hat d = \sum_{j \in J_\khat} d_j \leq d$.
\end{lemma}
We now have the main tools for proving \cref{thm:conf_int}.

\begin{proof}[\textbf{Proof of Theorem \ref{thm:conf_int}}]
\label{proof:conf_int}
Assume that $f \in \tH$, and that $\norm{f}_\tk \leq B$. Then by Theorem \ref{thm:RKHS_recovery} $f \in \Hhat$, with probability greater than $1-\delta$. Define $\hat B$ as
\begin{equation}\label{eq:def_bhat}
\hat B = B\left(1+\frac{\epsilon(n,m)}{2c_1}+ o\left(\epsilon(n,m)\right)\right),
\end{equation}
then by Lemma \ref{lem:norm_bound}, $\norm{f}_\khat \leq \hat B$ with probability greater than $1-\delta$. 
We first condition on the event that $f \in \Hhat$ and $\norm{f}_\khat \leq \hat B$. Then Lemma \ref{lem:vanilla_CI} gives the following confidence interval,
\begin{equation}
    \sP\left(\vert \hat\mu_{t-1}(\vx) - f(\vx)\vert \leq \hat\sigma_{t-1}(\vx)\left( \hat B + \sigma\sqrt{2(\hat \gamma_{t-1}+1+\log(1/\delta))}\right) \vert f \in \Hhat, \norm{f}_\khat \leq \hat B \right) \geq 1-\bar \delta
\end{equation}
We now remove the conditional event. Let $C_t(\vx):= \left[\hat\mu(\vx)-\hat\sigma(\vx), \hat\mu(\vx)-\hat\sigma(\vx)\right]$. By the chain rule,
\[
\sP\left(f(\vx) \in C_t(\vx)\right) \geq \sP\left(f(\vx) \in C_t(\vx)\vert f \in \Hhat\right)\cdot \sP\left(f \in \Hhat\right) \geq 1 - \delta - \bar\delta
\]
Renaming $\delta + \bar\delta$ to  $\delta$ for simplicity, we conclude that with probability greater than $1-\delta$
\[
\vert \hat\mu_{t-1}(\vx) - f(\vx)\vert \leq \hat\sigma_{t-1}(\vx)\left( B\left(1+\frac{\epsilon(n,m)}{2c_1}+ o\left(\epsilon(n,m)\right)\right) + \sigma\sqrt{2(\hat \gamma_{t-1}+1+\log(1/\delta))}\right).
\]
Lastly, Lemma \ref{lem:infogain} gives an upper bound for $\hat\gamma_{t-1}$ which completes the proof.
\end{proof}

\begin{proof}[\textbf{Proof of Lemma \ref{lem:infogain}}]
For this proof, we follow a similar technique as in \citet{vakili2021information}.
Recall that $\khat(\vx,\vx') = \sum_{j\in J_\khat}\etahat_j\vphi_j^T(\vx)\vphi_j(\vx')$, 
where $J_{\khat} = \{1\leq j\leq p: \etahat_j \neq 0 \}$. Define $\hat d$ the effective dimension of kernels $k_j$ that correspond to this index set,  $\hat d = \sum_{j \in J_\khat} d_j$. Now consider any arbitrary sequence of inputs $(\vx_\tau)_{\tau=1}^t$
and let $\hat \Phi_t = \left(\hat \vphi(\vx_1), \cdots, \hat \vphi(\vx_t)\right) \in \sR^{t\times \hat d}$, with $\hat \vphi(\vx) = \left(\vphi_j(\vx)\right)_{j \in J_\khat}$. Define the $\hat d\times \hat d$ matrix $\Lambda = \mathrm{diag} \left((\etahat_j)_{j \in J_\khat}\right)$ as the diagonal matrix containing the eigenfunctions of $\khat$. We have $K_t = \hat\Phi_t\Lambda\hat \Phi_t^T$. Let $H_t = \Lambda^{1/2} \hat\Phi^T \hat\Phi \Lambda^{1/2}$, by the Weinstein-Aronszajn identity,
\begin{equation*}
\begin{split}
    \frac{1}{2}\log\det(\mI + \bar\sigma^{-2}K_t) & = \frac{1}{2}\log\det(\mI + \bar\sigma^{-2}H_t)\\
    & \leq \frac{1}{2} \hat d \log\big( \text{tr}(\mI + \bar\sigma^{-2}H_t)/\hat d\big)
\end{split}    
\end{equation*}    
For positive definite matrices $\mP \in \mathbb{R}^{n\times n}$, we have $\log\det \mP \leq n \log \text{tr} (\mP/n)$. The inequality follows from  $\mI + \bar\sigma^{-2}H_t$ being positive definite, since $\etahat_j \geq 0$. We may write,
\begin{equation*}
\begin{split}
    \frac{1}{2}\log\det(\mI + \bar \sigma^{-2}K_t) & \leq \frac{1}{2} \hat d \log\big(  1 +  \frac{\bar\sigma^{-2}}{\hat d}\text{tr}(\Lambda^{1/2} \hat\Phi^T \hat \Phi \Lambda^{1/2}) \big) \\
    & \leq \frac{1}{2} \hat d \log\big(  1 +  \frac{\bar\sigma^{-2}}{\hat d}\sum_{\tau=1}^t\text{tr}(\Lambda^{1/2} \hat\vphi^T(\vx_\tau) \hat\vphi(\vx_\tau)  \Lambda^{1/2}) \big) \\
    & \leq \frac{1}{2} \hat d \log\big(  1 +  \frac{\bar\sigma^{-2}}{\hat d}\sum_{\tau=1}^t \vert\vert\hat\vphi({\vx_\tau})\Lambda^{1/2}\vert\vert_2^2 \big) \\ 
    &  \leq \frac{1}{2} \hat d \log\big(  1 +  \frac{\bar\sigma^{-2}}{\hat d}\sum_{\tau=1}^t \sum_{j \in J_\khat} \etahat_j \norm{\vphi_j(\vx_\tau)}_2^2 \big) \\ 
    &  \leq \frac{1}{2} \hat d \log\big(  1 +  \frac{\bar\sigma^{-2}t}{c_1}\max_j\norm{\vbetahat\gj}_2 \big)
\end{split}    
\end{equation*}
The next to last inequality holds since $k_j(\vx,\vx) = \norm{\vphi_j(\vx)}_2^2$ is normalized to one and $\etahat_j = \norm{\vbetahat_\gj}_2/c_1$. The inequality above holds for any sequence $(\vx_\tau)_{\tau\leq t}$,
\[
\hat\gamma_{t} = \gO\left(\hat d \log(t/c_1)\right)
\]
\end{proof}

\subsection{Proof of \cref{cor:regret_bound}}
Similar to the previous section, we take advantage of a classic regret bound for the oracle learner and then apply it to our setting. 

\begin{lemma}[Theorem 3 \citet{chowdhury2017kernelized}, for $\khat$]\label{lem:vanilla_regret}
Set $\delta \in (0,1)$. If $f \in \Hhat$ with $\norm{f}_\khat \leq \hat B$, then with probability $1-\delta$, \textsc{GP-UCB} satisfies,
\[
R_T = \gO\left(\hat B \sqrt{T\hat\gamma_T} + \sqrt{T\hat\gamma_T (\hat\gamma_T + \log1/\delta)}\right)
\]
\end{lemma}

\begin{proof}[\textbf{Proof of Corollary \ref{cor:regret_bound}}]\label{proof:regret_bound}
Conditioned on the event that $f \in \Hhat$ and $\norm{f}_\khat \leq \hat B$, Lemma \ref{lem:vanilla_regret} states that 
\[
    \sP\left(R_T = \gO\left(\hat B \sqrt{T\hat\gamma_T} + \sqrt{T\hat\gamma_T (\gamma_T + \log1/\delta)}\right) \vert f \in \Hhat, \norm{f}_\khat \leq \hat B \right) \geq 1-\bar \delta
\]
Then by Lemma \ref{lem:norm_bound} and Theorem \ref{thm:RKHS_recovery}, with probability greater than $1-\delta-\bar\delta$
\[
    R_T = \gO\left(\hat B \sqrt{T\hat\gamma_T} + \sqrt{T\hat\gamma_T (\hat\gamma_T + \log1/\delta)}\right) 
\]
where $\hat B$ is set according to \cref{eq:def_bhat}. Plugging in Lemma \ref{lem:infogain} to bound the information gain and changing the variable name $\delta + \bar\delta$ to $\delta$ and concludes the proof.

\end{proof}
\section{Experiments} \label{app:experiments}

\paragraph{Regret Experiment on Hyper-Parameter Tuning data}
This experiment is based on \citet[][Appendix B.1.2]{rothfuss2021meta}, we repeat some of the details for completeness. 
We consider the use case of hyper-parameter tuning for machine learning algorithms. In particular, we consider Generalized linear models with elastic NET regularization \textsc{(GLMNET)} \citep{friedman2010regularization} for this purpose, which has two hyper-parameters \texttt{lambda} and \texttt{alpha}. 
Following previous work \citep[e.g.,][]{perrone2018}, we replace the costly training and evaluation step by a cheap table lookup based on a large number of hyper-parameter evaluations \citep{kuhn2018automatic} on 38 classification datasets from the OpenML platform \citep{Bischl2017OpenMLBS}. The hyper-parameter evaluations are available under a Creative Commons BY 4.0 license and can be downloaded here\footnote{\href{https://doi.org/10.6084/m9.figshare.5882230.v2}{\texttt{https://doi.org/10.6084/m9.figshare.5882230.v2}}}.  In effect, $\gX$ is a finite set, corresponding to 10000-30000 random evaluations hyper-parameter evaluations per dataset and machine learning algorithm. Since the sampling is quite dense, for the purpose of empirically evaluating the meta-learned models towards BO, this finite domain can be treated like a continuous domain. All datasets correspond to binary classification. The target function we aim to optimize is the area under the ROC curve (AUROC) on a test split of the respective dataset.
Since \citep{kuhn2018automatic} sample \texttt{lambda} in the log-space, we transform it via $\texttt{lambda} \leftarrow \log_2(\texttt{lambda}) / 10$ such that we can expect a reasonably good performance of a Vanilla \textsc{GP-UCB} with SE kernel.
We randomly split the available tasks (i.e. train/test evaluations on a specific dataset) into a set of meta-train and meta-test tasks. In the following, we list the corresponding OpenML dataset identifiers:
\vspace{-10pt}
\begin{itemize}
\item meta-train tasks: 3, 1036, 1038, 1043, 1046, 151, 1176, 1049, 1050, 31, 1570, 37, 4134, 1063, 1067, 44, 1068,  50, 1461, 1462 \looseness -1
\vspace{-7pt}
\item test tasks: 335, 1489, 1486, 1494, 1504, 1120, 1510, 1479, 1480, 333, 1485, 1487, 334
\end{itemize}
\vspace{-10pt}
The plots in \cref{fig:regret_bo_2d} show average cumulative regret for 13 test tasks, and each tested for 10 runs with different random seeds.

\begin{figure}[ht]
    \centering
    \includegraphics[width = 0.5\linewidth]{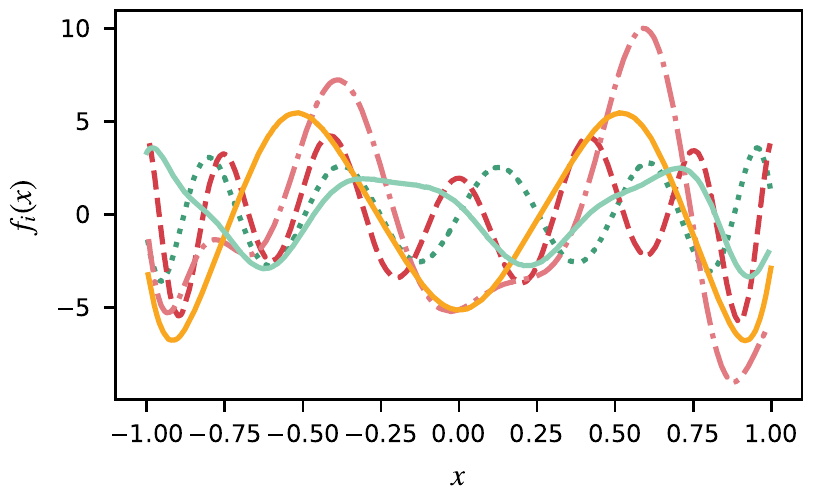}
    \caption{Examples of possible functions $f_s$ for the meta-dataset.}
    \label{fig:functions}
\end{figure}

\begin{figure}[ht]
    \centering
    \includegraphics[width = 0.4\linewidth]{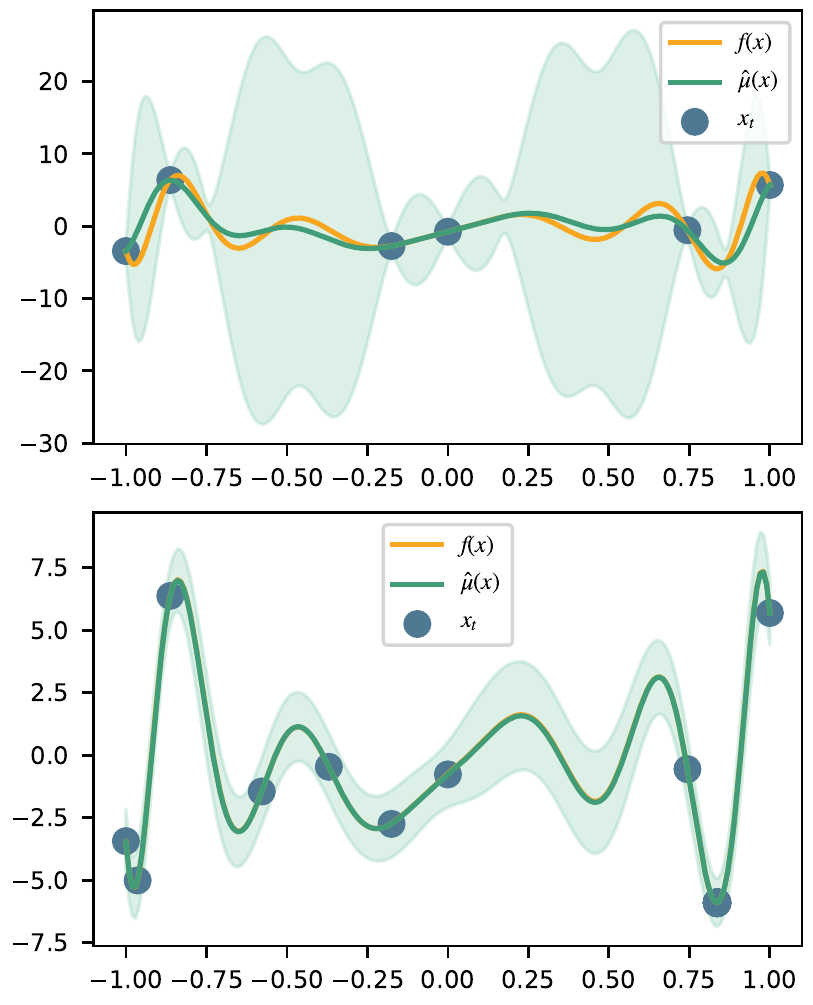}
    \caption{BO (minimization) with \mmkl. Upper plot shows the state at $t=5$ and the lower plot at $t=55$.}
    \label{fig:sample_bo}
\end{figure}

\begin{figure}[ht]
    \centering
    \includegraphics[width = 0.7\linewidth]{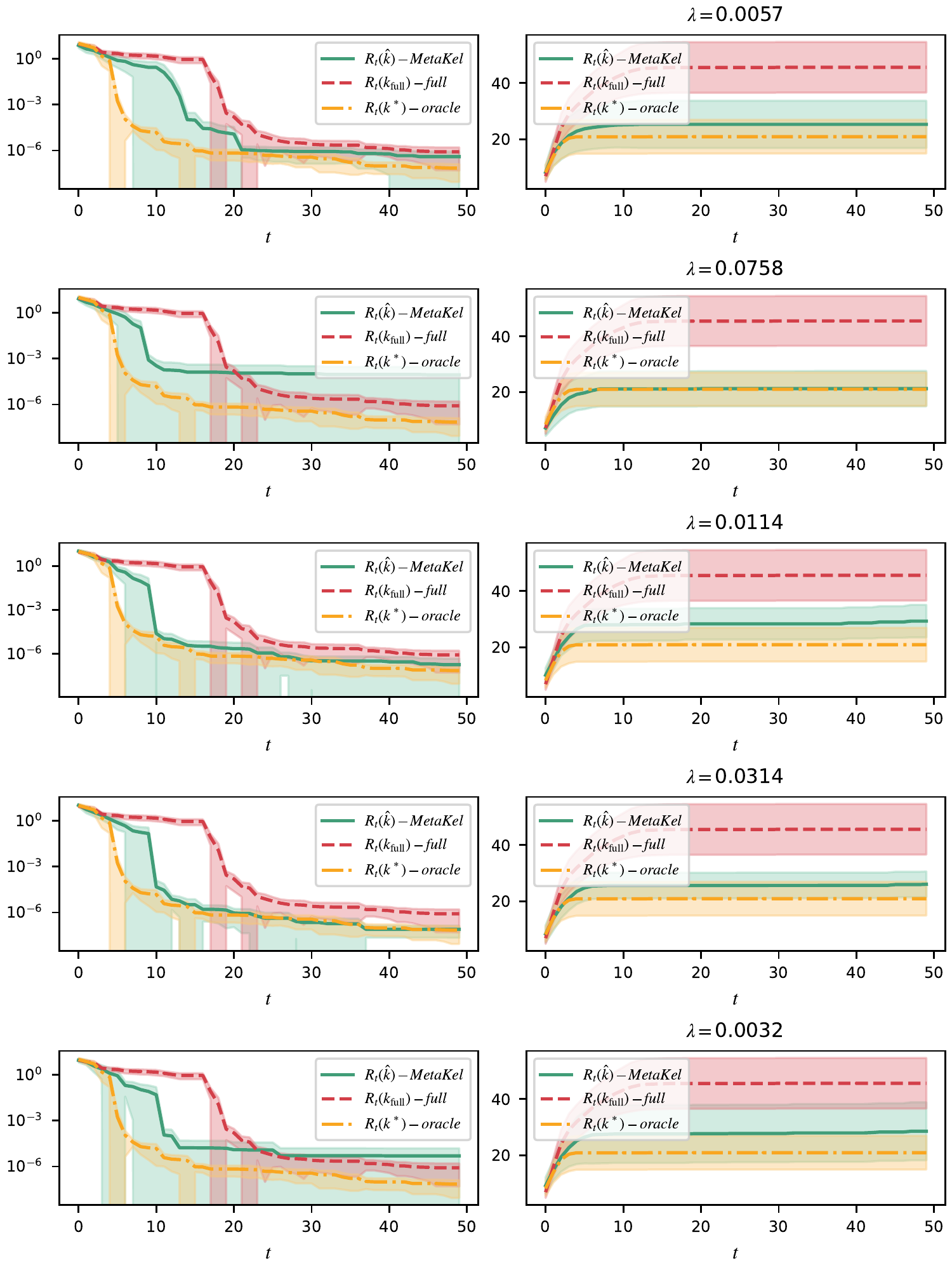}
    \caption{For $m=n=50$ and $p=20$, \cref{thm:RKHS_recovery} requires that $\lambda > 0.001$ for recovery to happen with probability greater than $1-\delta=0.9$. For $\lambda$ that satisfies this condition, the particular choice of its value does not effect performance severely.}
    \label{fig:lambda_choice}
\end{figure}

\paragraph{Supplementary Figures}
\cref{fig:functions} illustrates a few samples of the random functions that we optimize over in the experiments. The functions are constructed using the Legendre basis, as explained in the main text.
\cref{fig:sample_bo} gives an example of a BO problem where we use \textsc{IGP-UCB} together with the meta-learned kernel to find the minimum of a function with as few samples as possible. As the function estimate improves, the confidence sets rapidly shrink and the learner only samples points close to the minimum.
\cref{fig:lambda_choice} demonstrates that the choice of $\lambda$ for the \mmkl loss does not have a severe effect on the regret of \textsc{IGP-UCB}. This is only the case if $\lambda$ satisfies the condition of \cref{thm:RKHS_recovery}.

\end{document}